\documentclass{report}

\usepackage{blindtext}
\usepackage[T1]{fontenc}
\usepackage[utf8]{inputenc}
\usepackage{graphicx}
\usepackage{comment}
\usepackage{amsmath,amssymb}
\usepackage{color}
\usepackage{url}
\usepackage{hyperref}

\usepackage{graphicx}
\usepackage{amsmath}
 \usepackage{nicematrix} % vertical text
\usepackage{multirow}
\usepackage{comment}
\usepackage{float} % to put figure H

\usepackage{listings}
\usepackage{xcolor}
\usepackage{tikz, fix-cm}
\usetikzlibrary{plotmarks}
\usepackage{graphicx,booktabs,array}

\usepackage{amssymb}% http://ctan.org/pkg/amssymb
\usepackage{pifont}% http://ctan.org/pkg/pifont
\def\secref#1{Sec.~\ref{#1}}
\def\chapref#1{Chap.~\ref{#1}}
\def\figref#1{Fig.~\ref{#1}}
\def\tabref#1{Tab.~\ref{#1}}
\def\eqref#1{Eq.~(\ref{#1})}

\newcommand\etal{\emph{et al.}}

\def\argmin{\mathop{\rm argmin}}

\begin{document}

\begin{titlepage}
   \begin{center}
       \vspace*{1cm}
        \begin{LARGE}  
       \textbf{Registration Techniques for Deformable Objects}
        \end{LARGE}
        
       \vspace{1.5cm}
        \begin{large} 
       \textbf{Alireza Ahmadi}
       
       \vspace{0.2cm}
       
       Zanjan, Iran

       \vspace{0.8cm}
     
       Supervisor: Prof.~Dr.~Cyrill Stachniss
       
       \vspace{0.8cm}

       \textbf{Institute of Photogrammetry and Robotics}
       
       Geodetic Engineering and Mobile Sensing
       
       University of Bonn
       
       Germany
       
       2020
       \end{large}
            
   \end{center}
\end{titlepage}

\tableofcontents{}
\cleardoublepage{}

\chapter{Introduction}

\section{Motivation} 
\label{sec:motivation}

The problem of mapping unknown objects and localizing the robot in environment is known as SLAM. This subject has become one of the most interesting fields in the recent years. Using SLAM methods, a robotic agent can move in the environment while localizing itself and building a map of the surroundings. This capability helps the robot to execute diverse tasks like autonomous navigation and manipulating objects in the environment.

The dense environment mapping has been studied a lot and the majority of the research focuses in the area of rigid mapping. 
The 3D reconstruction can be done through matching consecutive scans taken over time in the environment \cite{izadi2011kinectfusion}. 
Most of the mapping algorithms assume the environment to be static, as a consequence, they fail in handling non-rigid models. By relaxing the rigidity assumption, the registration task gets more challenging. Scans of an object which undergo deformations, can not be registered with a naive rigid transformation anymore. In such cases, the algorithm should be able to estimate deformation parameters to match the source scan to the target optimally \cite{izadi2011kinectfusion}. The deformation parameters can be estimated via deformation graph method. This method sub-samples the scanned surface of deforming object by a set of 3D points. Then, for each pair of correspondent points, we find the best deformation parameters which can retrieve the original shape of the deforming object.
Acquisition of surface deformation parameters is a necessary ingredient of mapping in dynamic environments. Estimating such surface and deformation parameters makes this task more challenging. The most critical task in this scope is to find the parameters which describe the best match between two consecutive scans while taking into account the non-rigidity of the target object. Also, at the same time, the pipeline should be able to keep the track of camera pose in the environment and be able to warp the deformations smoothly and efficiently.

\section{Problem Statement} 
\label{sec:problem_statement}

In general, the problem of non-rigid registration is about matching two different scans of a dynamic object taken at two different points in time. These scans can undergo both rigid motions and non-rigid deformations. Since new parts of the model may come into view and other parts get occluded in between two scans, the region of overlap is a subset of both scans. In the most general setting, no prior template shape is given and no markers or explicit feature point correspondences are available. So, this case is a partial matching problem which takes into account the assumption that consequent scans undergo small deformations while having a significant amount of overlapping area \cite{li2008global}. The problem which this thesis is addressing is about mapping deforming objects and localizing camera in the environment at the same time.

\section{Main Contributions} % (fold)
\label{sec:main_contributions}

%What the goal of your thesis? This section is for in-depth overview of the
%contributions you have made in the thesis. Dedicate a paragraph to a single
%contribution if you have many and elaborate on what is so special about any of
%these.
In this thesis, we contribute to the subject of the SLAM problem, where a robotic agent navigates in the environment and constructs a dense 3D map of the surrounding static and dynamic objects. In this section, we describe the main contributions of this thesis as follows:

\begin{itemize}
	\item The first contribution of this thesis is an approach for constructing a dense representation of the environment while accurately tracking camera pose. The proposed technique is based on $KinectFusion$ \cite{izadi2011kinectfusion} and voxel hashing technique \cite{niessner2013real}. 
	Our approach uses geometric and photometric information of the scene to estimate the camera pose precisely. The performance of our approach is evaluated via different data-sets of static scenes from TUM and NUIM collections, showing the robustness of our approach in handling harsh movements and dealing with outliers. The proposed method is explained in details in \chapref{chap:dense_rigid_slam}.
	
	\item The second contribution of this work is an approach that enables robotic agents to map objects deformations. The proposed method can model deformations of a non-rigid object without using any pre-build model or any explicit template. Our method is inspired from deformation graph idea ~\cite{sumner2007embedded} and $DynamicFusion$ ~\cite{newcombe2015dynamicfusion}, which enables us to map non-rigid scenes.
\end{itemize}

In sum, this thesis presents two contributions, that address different issues
in the context of dense 3D mapping. All the methods are designed to work online on real-world data. Also, all algorithms are developed using C++ and CUDA libraries to reach real-time performance. The open-source code of our implementation is available at: 

\url{https://github.com/alirezaahmadi/dynamap}

\section{Thesis Overview} % (fold)
\label{sec:thesis_overview}

In this thesis, we propose a method for mapping rigid and non-rigid objects using RGB-D sensors. In \chapref{chap:basic_techniques} we introduce the basic techniques which are used in our method. In \chapref{chap:dense_rigid_slam}, we explain the necessary components of building a dense rigid SLAM system, including theoretical roots and derivation of normal equations. The \chapref{chap:dense_non_rigid_slam} includes details about introducing non-rigidity into the SLAM problem and discusses crucial aspects in dealing with deforming objects in the context of SLAM system. This includes methods of modeling non-rigidities, building data associations, estimation of deformations and warping deformation techniques.

%Use this section only if it is necessary for your thesis.
\clearpage
\cleardoublepage{}

\chapter{Basic Techniques}
\label{chap:basic_techniques}
% !TEX root = ../thesis.tex

%\lettrine{T}{his} section is dedicated to 

%*********************************************************************************************
 Our main goal in this work is to generate dense 3D reconstructions of rigid and non-rigid objects. In this chapter, we introduce the basic techniques required for accomplishing the task. These include several concepts from computer vision literature, non-linear least-squares optimization techniques and other data management tools required for efficient computation. 
%*********************************************************************************************

\section{Input Data}
\label{sec:InputData}
\begin{figure}[h]
	\centering 
	\includegraphics[width=0.35\linewidth]{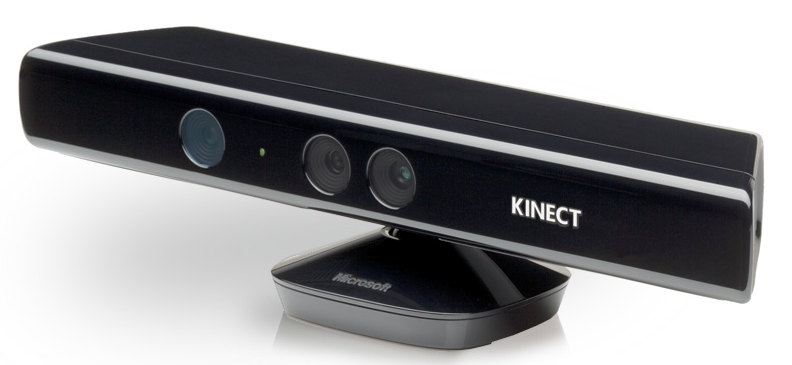}
	\caption[RGB-D commodity sensor]{RGB-D commodity sensors, Microsoft Kinect}
	\label{fig:sensor}
\end{figure}

High-resolution dense mapping of the environment became viable through the use of relatively new RGB-D commodity cameras like Microsoft Kinect sensor shown in \figref{fig:sensor}. These devices provide real-time high-resolution depth and RGB images. Using these cameras eases the process of data collection for SLAM systems. However, these sensors suffer from some inevitable problems like disturbances due to presence of noise or outliers and the instability of depth maps in extreme light conditions. 

\begin{figure}[h]
	\centering
	\includegraphics[width=0.4\linewidth]{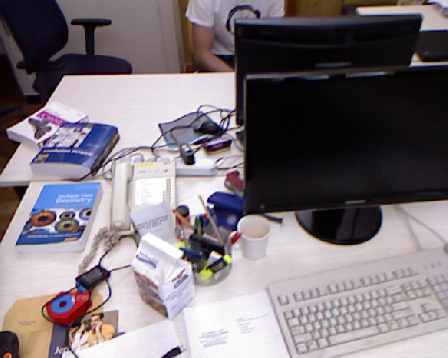}
	\includegraphics[width=0.4\linewidth]{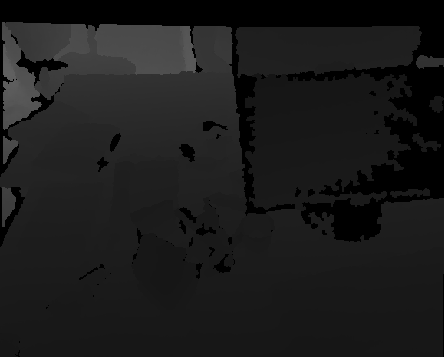}
	\caption[RGD and Depth Images]{RGD and Depth Images, $right$: Depth image in Gray-scale visualization, $left$: regular RGB image. }
	\label{fig:depthRGB}
\end{figure}

The depth values are visualized as intensity image with pixel values between 0-255 which represent the distance of corresponding point in the real-scene to camera center. The actual depth measurements (in meters) can be computed from these intensity values using the camera model. Also, RGB images as their name yields has three different channels (red, blue and green) to represent the color captured from the point in the real environment. An example of the depth and RGB image from a Kinect sensor is shown in \figref{fig:depthRGB}.

%********************************************************************************************
\section{Camera Model}
\label{sec:camera_model}

\begin{figure}[h]
	\centering
	\includegraphics[width=0.7\linewidth]{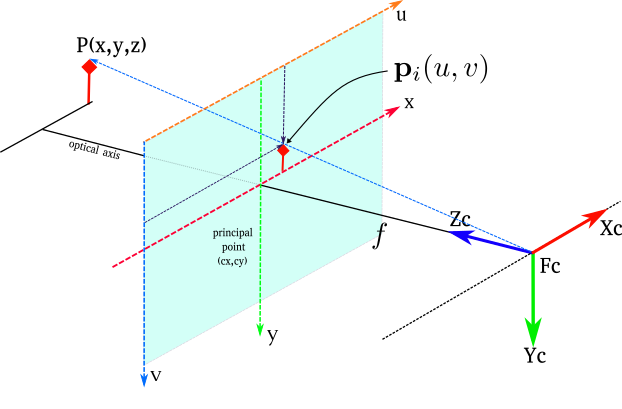}
	\caption[Pinhole camera model]{Pinhole camera model, showing 3D point $P$ which is projected into 
		2D space of image plane with pixel coordinates $p$.}
	\label{fig:projectivecamera}
\end{figure}

The camera model is a mathematical description of the imaging process via a projection function for a camera or vision sensor. The projection function defines how a 3D point is projected into the image plane of the camera given its intrinsics. In our implementation, we use the pinhole projection model \cite{sturm2014pinhole}. We show the main elements of the model in \figref{fig:projectivecamera}.

In \figref{fig:projectivecamera}, a camera with center of projection denoted with $O$ and the principal axis parallel to the $Z_c$ axis going through image plane are shown. The image plane is placed along the $Z_c$ axis at a distance equal to the focal length $f$ away from center of the projection $O$. Using this model, a 3D point $\mathbf{P}(x, y, z)$ will project into the image plane forming point $\mathbf{p}(u, v)$, according to equation below:

\begin{equation}
\label{eq:projectivecameraS}
\mathbf{p}(u, v) = \pi(K\mathbf{P})
\end{equation}

where $\pi(w)$ represented the homogeneous normalization and the calibration matrix $\mathbf{K}$ defines as follows:
 
\begin{equation}
\label{eq:cameraK}
\begin{split}
\mathbf{K} &= 
\begin{pmatrix}
f_x & 0 & c_x\\
0 & f_y & c_y\\
0 & 0 & 1
\end{pmatrix} \\
\pi(v) &= \dfrac{1}{v_z} 
\begin{pmatrix}
v_x\\
v_y\\
1
\end{pmatrix}
\end{split}
\end{equation}
with $f_x, f_y$ as focal lengths in x and y direction and the principal 
point $c_x , c_y$ respectively in x and y axis (in pixels) \cite{hartley2005multiple}.
%*********************************************************************************************
\subsection{Surface Normals}
\label{subsec:surface_normals}

Surface normals are vectors perpendicular to the surface of objects. They are useful to identify geometric attributes like the orientation and the planarity of the objects in a scene. In our work, we use the normal information for aligning point clouds and for outlier rejection during the correspondence estimation step.
%  To estimate normals from the depth image, we use \ref{eq:plane} as plane equation, where $s$ and $t$ are real numbers and two vectors $\mathbf{v}_1$ and $\mathbf{v}_2$ are vectors crossing from the point $\mathbf{q}_0$ which is fixed on the plane $\mathbf{q}$.

\begin{figure}[h]
	\centering
	\includegraphics[width=0.8\linewidth]{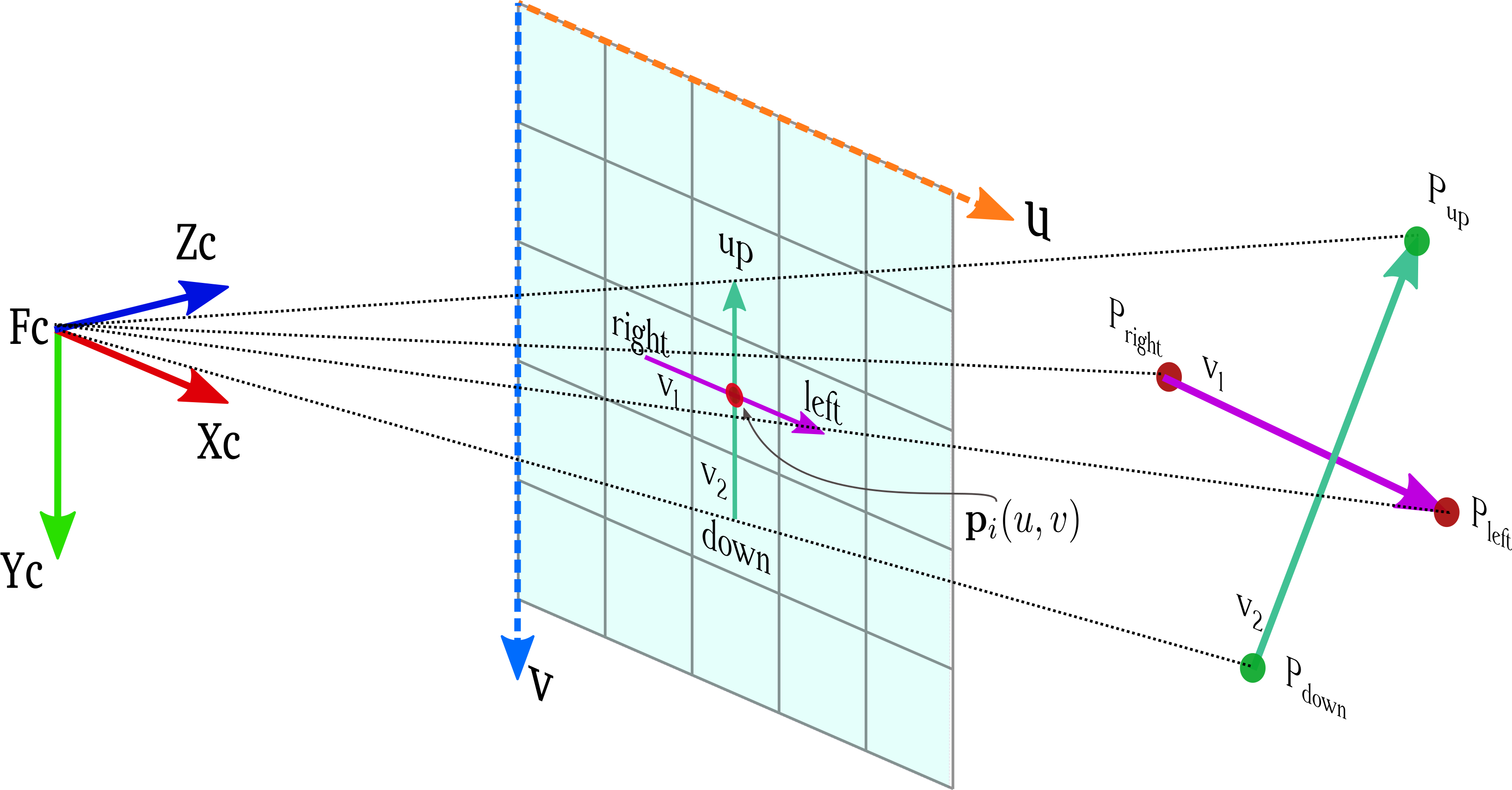}
	\caption[Normals from depth image]{Two vectors $\mathbf{v}_1, \mathbf{v}_2$ are constructed based on depth values of neighbor pixels of target pixel $\mathbf{p}_i$. Using $v_1$ and $v_2$ normal at pixel $\mathbf{p}_i$ can be estimated.}
	\label{fig:surface_normals}
\end{figure}

% \begin{equation}
% \label{eq:plane}
% \mathbf{q} = \mathbf{q}_0 + s \mathbf{v}_1 + t \mathbf{v}_2
% \end{equation}

\begin{figure}[h]
	\centering
	\includegraphics[width=0.6\linewidth]{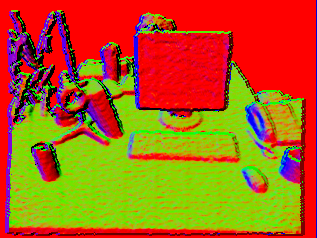}
	\caption[Surface normals result]{A color coded representation of surface normals is shown where colors blue, green and red show normals in directions $x, y$ and $z$ respectively.}
	\label{fig:result_surface_normals}
\end{figure}

To estimate normal at pixel location $\mathbf{p}_i$ in the depth image, we pick two neighboring pixels in the $u$ and $v$ directions from a window.
Then we compute the 3D points corresponding to these pixels and construct the vectors $\mathbf{v}_1$ and $\mathbf{v}_2$. Given these vectors, we compute the normal using the cross product operation. The size of the window that neighbors are selected from typically changes w.r.t the resolution of the image. In practice, we found the optimum range is between 1 two 5 pixels away from center $\mathbf{p}_i$. In \figref{fig:result_surface_normals} result of our implementation is shown.

\begin{equation}
\label{eq:normal_plane}
normal(\mathbf{p}_i) = \mathbf{v}_1 \times \mathbf{v}_2 
\end{equation}

%*********************************************************************************************
\subsection{Bilateral Filter}
\label{subsec:bilateral_filter}

The bilateral filter is a non-iterative smoothing filter which de-noises the pixels in the image while preserving edges by means of a non-linear combination of nearby image intensity values. 

\begin{figure}[h]
	\centering
	\includegraphics[width=0.9\linewidth]{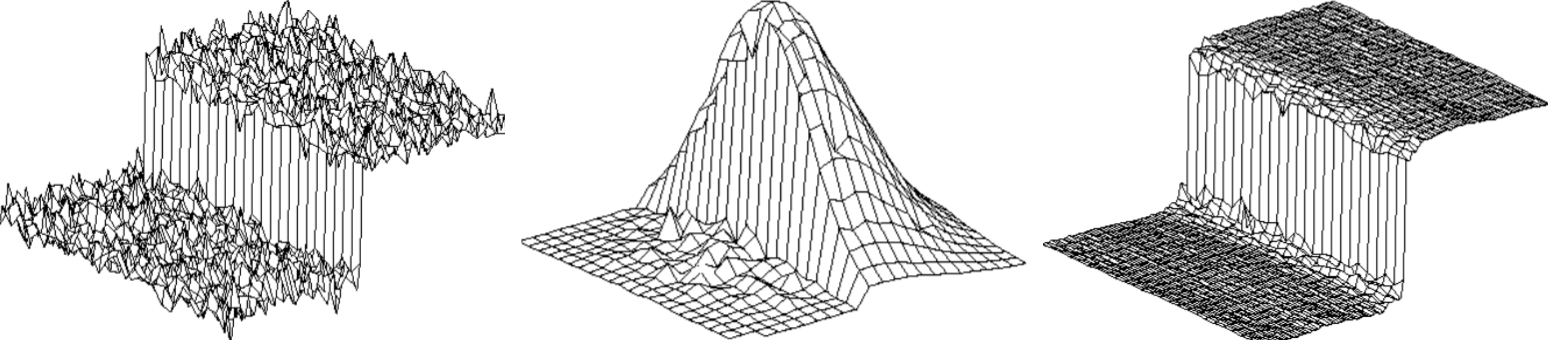}
	\caption[Bilateral filter]{\textit{Left:} A function containing a sharp edge and corrupted with random noise. \textit{Middle:} Bilateral filter kernel. \textit{Right:} Filtered function output preserving edge and properties of main function \cite{tomasi1998bilateral}.}
	\label{fig:bilateral_filter_pic}
\end{figure}

The bilateral filter smooths the input image by considering both the geometric closeness (i.e. how close are the pixels?) and photometric similarity (i.e. how similar is the intensity of the pixels) of nearby pixels. This is different from other smoothing filters which typically consider only the 
geometric closeness of the pixels. As a result, the bilateral filter is able to preserve the edges by exploiting the photometric closeness while smoothing away the noise \cite{tomasi1998bilateral}. The filter kernel $k(x)$ is defined as:

\begin{equation}
\label{eq:bilateralfilter_eq}
k(\mathbf{p}_i) = \mathbf{\int_{-\infty}^{\infty} \int_{-\infty}^{\infty}} c(\xi - \mathbf{p}_i)s(\mathbf{f(\xi)} - \mathbf{p}_i)d\xi 
\end{equation}

where $c$ measures the geometric closeness of center pixel $\mathbf{p}_i$ to its neighbor $\xi$ and $s$ measures photometric closeness.

An important case of bilateral filtering is the shift-invariant Gaussian filtering, in which both the closeness function $c$ and the similarity function $s$ are radially symmetric Gaussian functions with $\sigma_s$ and $\sigma_c$ as the respective standard deviations. These are defined as: 

\begin{equation}
\label{eq:bilateralfilter_s_eq}
s(\xi - \mathbf{p}_i) = e^{-\dfrac{1}{2} \left( \dfrac{d(\xi -\mathbf{p})}{\sigma_s} \right) ^2}
\end{equation}

\begin{equation}
\label{eq:bilateralfilter_c_eq}
c(\xi - \mathbf{p}_i) = e^{-\dfrac{1}{2} \left( \dfrac{\delta(\textbf f(\xi) - \textbf f(\mathbf{p})}{\sigma_c} \right) ^2}
\end{equation}

where

\begin{equation}
\label{eq:bilateralfilter_delta_eq}
\delta(\textbf f(\xi) - \textbf f(\mathbf{p}_i)) = \big| \textbf f(\xi) - \textbf f(\mathbf{p}_i) \big|
\end{equation}

where $d(\xi - \mathbf{p})$ and $\delta(\xi - \mathbf{p})$ denote the geometric and photometric differences between two pixels $\xi$ and $\mathbf{p}$ in the image, respectively. 

%*********************************************************************************************
\subsection{Image Pyramid}
\label{subsec:image_pyramid}

\begin{figure}[h]
	\centering
	\includegraphics[width=0.6\linewidth]{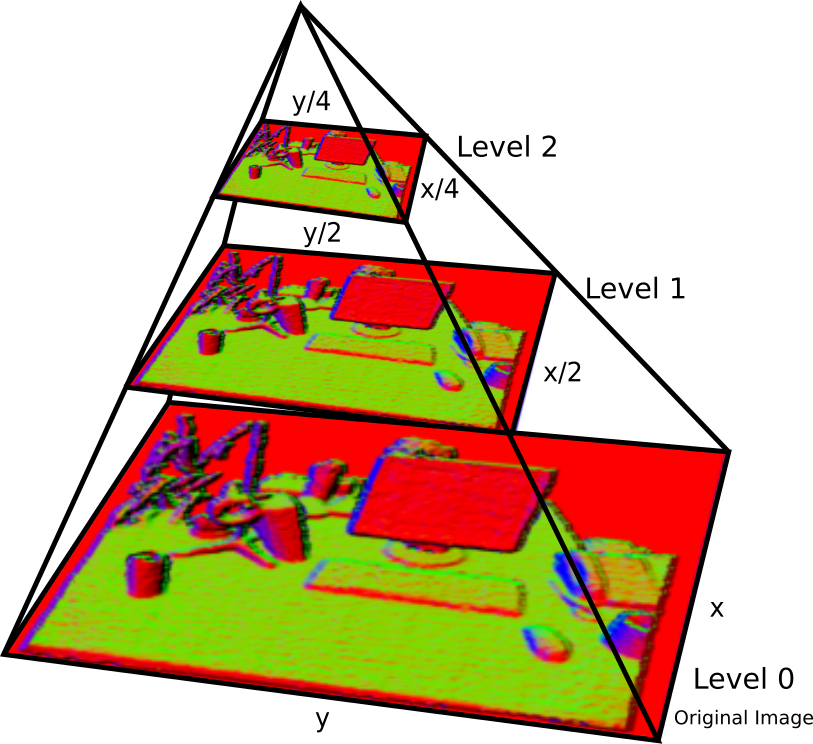}
	\caption[Multi-resolution Pyramid structure]{A multi-resolution pyramid structure containing two higher level of down-sampled images are shown. The original image forms the based of the pyramid at $level_0$.}
	\label{fig:pyramid}
\end{figure}

The pyramid technique scales the images while maintaining their main features. The motivation behind the pyramid technique is that the surrounding pixels within a certain area often have similar characteristics, and thus, they are highly correlated with each other, in such a way that by removing them the main content won't be damaged. This technique places the original image at the first level of a hypothetical pyramid and adds down-scaled images at the higher levels as illustrated in \figref{fig:pyramid}. We use this method within the course-to-fine procedure \secref{subsec:coarse_to_fine_processing}. Using this technique and iterating on different scales of the original data, the alignment procedure gets faster than directly working on the original image. 

To have a smooth scaled image, we first smooth the original image with an appropriate filter kernel (like Gaussian filter \secref{subsec:gaussian_filter}) using the convolution operation and then sub-sample the smoothed image.  The sub-sampling is done by eliminating odd rows and columns from the image which results an image with the half size of the original image size.
As mentioned before in order to improve computational efficiency and convergence while estimating the camera pose in \secref{sec:camera_pose_estimation}, a multi-resolution pyramid is defined \eqref{eq:warp_pyramid} at pixel $\mathbf{p}_i^l$. 

\begin{equation}
\label{eq:warp_pyramid}
\textbf{I}_i^{l+1} = \textbf{I}_i^l(w(\textbf{H}^l, \textbf{p}_i^l))
\end{equation}

where $\textbf{I}_i^{l+1}$ shows the image at higher level and $\textbf{H}^l$ is a matrix effecting the camera intrinsics as:

\begin{equation}
\label{eq:warp_pyramid_k}
\mathbf{H}^l = 
\begin{bmatrix} 
\lambda & 0 & 0 \\ 
0 & \lambda & 0 \\ 
0 & 0 & 1  
\end{bmatrix} \mathbf{K} \in \mathcal{R}^{3\times 3}
\end{equation}

where, $K$ is the camera calibration matrix and $\lambda = 1/2^l$ denoting the scale ratio.
%\textcolor{red}{What is $P_i$ in \eqref{eq:warp_pyramid}?}
%*********************************************************************************************
\subsection{Gaussian Filter}
\label{subsec:gaussian_filter}

Gaussian blur is a widely used method in computer vision to reduce image noise and remove details from the image before detecting relevant edges. Gaussian blur is a low-pass filter, reducing high frequency components of the image \cite{adelson1984pyramid}. In the smoothing procedure, a kernel like the one shown in \eqref{eq:gaussian_kernel} is used to weigh and smooth the image. 

\begin{equation}
\label{eq:gaussian_kernel}
H=\frac{1}{256}\left(\begin{array}{ccccc}1& 4& 6& 4& 1\\ {}4& 16& 24& 16& 4\\ {}6& 24& 36& 24& 6\\ {}4& 16& 24& 16& 4\\ {}1& 4& 6& 4& 1\end{array}\right)
\end{equation}

%*********************************************************************************************
\section{Non-Linear Least Squares Optimization}
\label{sec:non_linear_least_squares}
Non-linear least-squares optimization is an unconstrained optimization techniques that fits a set of $m$ observations with a model that can be expressed with $n$ unknowns non-linearly, where $m \geq n$. The general from of this optimization method is shown in \eqref{eq:least_squares_from}

\begin{equation}
\label{eq:least_squares_from}
\textbf{F(\textbf{x})} = \dfrac{1}{2} \sum_{i=1}^{m} \mathbf{f}_i(\mathbf{x})^2 =\dfrac{1}{2} \mathbf{f}(\textbf{x})^\top \mathbf{f}(\textbf{x})
\end{equation}

where $\mathbf{f}_i(x)$ is the difference between one of the desired and predicted values in the system and objective function $\textbf{F(\textbf{x})}$ is defined based on the sum of the differences of the squares. A typical way to minimize this function is to find the local minimum of the function and iteratively update the estimates in direction of descent \cite{madsen1999methods}.

\subsection{Gauss-Newton method}
\label{subsec:gauss_newton_method} 

While there are a bunch of different methods to minimize function \eqref{eq:least_squares_from}, we use the Gauss-Newton method, that finds $x^* = \argmin_x \textbf{F(\textbf{x})}$ by taking first derivative of $\mathbf{F(x)}$ using Taylor expansion for small steps $\mathbf{h}$ where:

\begin{equation}
\label{eq:linearlize_ls}
\mathbf{f}(\mathbf{x} + \textbf{h}) \cong \ell(\textbf{h}) \equiv \mathbf{f}(\mathbf{x}) + \textbf{J}(\mathbf{x}) \textbf{h}
\end{equation}

This method approximates $\mathbf{f}$ by linearizing it around $x$ and assumes the function $\mathbf{f}$ to be locally quadratic. By inserting \eqref{eq:linearlize_ls} into \eqref{eq:least_squares_from} we get:

\begin{equation}
\begin{split}
\mathbf{F}(\textbf{x} + \textbf{h}) \cong L(\textbf{h})  & \equiv \dfrac{1}{2} \ell(\textbf{h})^\top \ell(\textbf{h}) \\
& = \dfrac{1}{2} \textbf{f}^\top \textbf{f} + \textbf{h}^\top \textbf{J}^\top \textbf{f} + \dfrac{1}{2} \textbf{h}^\top \textbf{J}^\top \textbf{J} \textbf{h} \\
& = F(\textbf{x}) + \textbf{h}^\top \textbf{J}^\top \textbf{f} + \dfrac{1}{2} \textbf{h}^\top \textbf{J}^\top \textbf{J} \textbf{h}
\end{split} 
\end{equation}

Given the Jacobian matrices $\textbf{J}$, we can now solve for minimizing the error which this procedure iteratively reduces the sum of squared errors toward the minimum of quadratic function $\mathbf{f}$ with steps of size of $\textbf{h}$.

\begin{equation}
(\textbf{J}^\top \textbf{J}) \textbf{h} = -\textbf{J}^\top \mathbf{r}
\end{equation}

where, $\mathbf{r}$ is know as residuals and shows actual error value at given point $\textbf{x}$. To update the value of $\textbf{x}$  \eqref{eq:update_x} is used ~\cite{madsen1999methods}.

\begin{equation}
\label{eq:update_x}
\textbf{x}_{t+1} = \textbf{x}_{t-1} + \textbf{h}
\end{equation}

\subsection{Levenberg-Marquardt method}
\label{subsec:levenberg_marquardt_method}

This method damps down the Gauss-Newton operations via a damping scalar $\alpha$ where objective equation would be defined as:
 
\begin{equation}
(\textbf{J}^\top \textbf{J} + \alpha \textbf{I}) \textbf{h} = -\textbf{J}^\top \mathbf{r}
\end{equation}

where, $ \textbf{I}$ is an identity matrix of size $n \times n$ and this extra coefficient ensures to always have positive definite $\textbf{J}^\top \textbf{J}$ as long as $\alpha > 0 $. In other words, this coefficient forces optimization to always take steps toward down-hill of the function even in cases where there is some rank deficiency issues. The \eqref{eq:alphe_LM} shows an approach to initialize the $\alpha$.

\begin{equation}
\label{eq:alphe_LM}
\alpha_0 = \tau \max_i(\textbf{J}^\top \textbf{J})
\end{equation}

where $\tau$ is a user defined scalar to ensure proper convergence. Its better to have bigger $\alpha$ at the beginning and decrease it over the time via the number of iterations. This way a better convergence is expected in comparison with Gauss-Newton method in situations with in-proper initial guess ~\cite{madsen1999methods}.

%*********************************************************************************************
\subsection{Robust Kernel}
\label{subsec:robust_kernel}

Even a few outliers present in the data can totally spoil an ordinary least squares solution. To cope with such challenges, statistical tools have been developed which helps to robustify the estimation procedure to outliers. A robust solution can acquired by using a re-weigthed least squares approach which tries to reduce the effect of outliers \cite{deleast} by assigning them a smaller weight during the optimization process.

\subsubsection{Huber Method}
\label{subsubsec:hubers_mthod}
The Huber kernel ~\cite{huber1992robust} defines a penalty based on the value of the residual $\mathbf{r}$. This method is used in our implementation to weight wrong correspondences through camera pose tracking optimization procedure. This robust function deals quadratically with small values of $(\mathbf{r} < \delta$) and linearly with larger values. The $\mathbf{r}$ typically called residuals where it defines based on the difference between predicted value and observations $\mathbf{r} = y(x) - \textbf{f}(x)$. The Huber weight function is defined as:

%\begin{equation}
%\phi(x)=\begin{cases}
%\dfrac{1}{2} r^2, & \text{if } \lvert r \rvert \leq \delta , \\
%\delta ( \lvert r \rvert - \dfrac{1}{2} \delta), & \text{otherwise}.
%\end{cases}
%\end{equation}

\begin{equation}
w(x)=\begin{cases}
1 , & \text{if } \lvert \mathbf{r} \rvert \leq \delta , \\
\delta / \lvert \mathbf{r} \rvert, & \text{otherwise}.
\end{cases}
\end{equation}

\subsubsection{Tukey Method}
\label{subsubsec:l1_mthod}

The Tukey method offered same structure, while its function zeros every instance with error greater than the threshold. Also, the function itself is not differentiable yet still can handle outliers very well.
In this work we use Tukey function in non-rigid optimization process to weight inconsistent correspondences. The equations below show definition of Tukey wight function based on $\lambda$ denoting user-defined control called Tukey coefficient ~\cite{tukey1979robust}.

%\begin{equation}
%\psi(x)=\begin{cases}
%\dfrac{\lambda ^ 2}{6} [1 - (1 - (\dfrac{r}{\lambda})^2)^3], & \text{if } \lvert r \rvert \leq \lambda , \\
%\delta ( \lvert r \rvert - \dfrac{1}{2} \delta), & \text{otherwise}.
%\end{cases}
%\end{equation}

\begin{equation}
w(x)=\begin{cases}
[1-(\mathbf{r}/\lambda)^2]^2 & \text{if } \lvert \mathbf{r} \rvert \leq \lambda , \\
0, & \text{otherwise}.
\end{cases}
\end{equation} 

%********************************************************************************************
\section{Dual-Quaternions}
\label{sec:dual_quaternions}
Non-rigid modeling involves warping deformations on surfaces efficiently and accurately. In computer graphics, there are a bunch of methods that can be used to warp the deformations on various surfaces. One of these methods is based on the concept on dual-quaternions. The dual-quaternion as its name suggests is a combination of two quaternions. The dual-quaternions can express a rigid transformation more concisely ~\cite{guo2012task}. We use Dual-Quaternions as long as, non-rigid mapping process involves using blending techniques to warp the deformations of the model.
Dual-quaternions have similar properties to quaternions such as being an un-ambiguous and singularity-free representation. In comparison with quaternions which can only represent rotation, the dual-quaternions can represent a rotation along with a translation using eight parameters. These parameters are divided into two parts: real and dual, which is denoted in form of \eqref{eq:dual_quat}.

\begin{equation}
\label{eq:dual_quat}
dq = \mathbf{q}_r + \mathbf{q}_d \xi
\end{equation}

where $q_r$ and $q_d$ represent the real and dual quaternions and $\xi$ is the dual-number part. Also, dual-quaternion has a unified representation of translation and rotation as follows:

\begin{equation}
\label{eq:dual_quatRT}
\begin{split}
\mathbf{q}_r &= \mathbf{r} \\
\mathbf{q}_d &= \dfrac{1}{2}\mathbf{t}.\mathbf{r}
\end{split}
\end{equation}

where $\mathbf{r}$ is a unit quaternion representing the rotation and the translation vector is represented by the vector $\mathbf{t} = \{x, y, z \}$.

\subsection{Dual-Quaternion Linear Blending}
\label{sec:dual_quat_linear_blending}
In the context of non-rgid registration, we need a method to apply estimated deformations on the reconstructed 3D model. We used Dual-Quaternion Linear Blending (DQLB) which is works based on linear combination of Dual-Quaternions. The traditional deformation blending pipelines assume a 3D model in a rest-pose with a set of joints $v$, and corresponding weight parameters $w$. The weight $w_i$ defines how much neighbors of $i_{th}$ joint will be effected by its motions. Each joint $v_i$ is associated with a transformation matrix $C_i$ which represents its pose and motion. The most popular technique performs blending via linearly combination of neighbor nodes transformation using \eqref{eq:lb} ~\cite{kavan2007skinning}.

\begin{equation}
\label{eq:lb}
v' = \sum_{i=1}^{n}w_i C_j v
\end{equation}

\begin{figure}[h]
	\centering
	\includegraphics[width=0.4\linewidth]{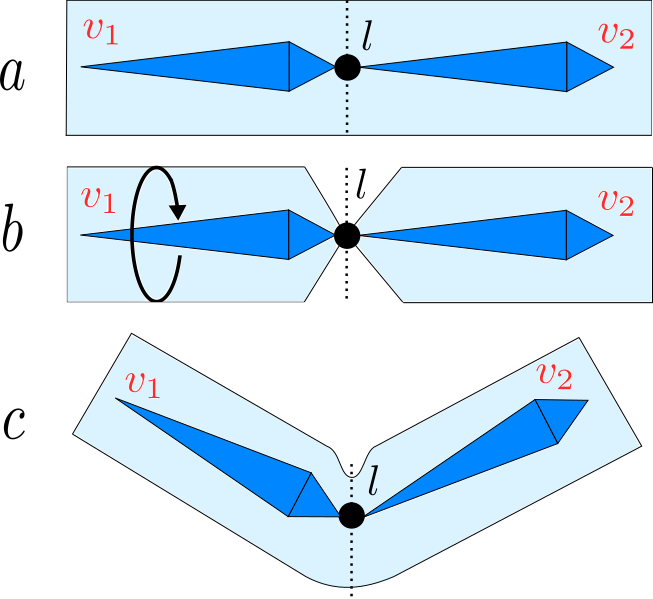}
	\caption[Linear blending artifacts]{Linear blending artifacts, in $a$: a model with two joints connected via link $l$ are shown in the rest-pose. $b$: shows situation where joint $v_1$ is rotated and caused volume-lose artifact in the model, and $c$ represents situation which $v_1$ and $v_2$ have rotational motion and blender failed to warp the deformations correctly. }
	\label{fig:lb_vloss}
\end{figure}

The linear blending suffers from skin collapsing and volume-loss artifacts \figref{fig:lb_vloss} because the result of \eqref{eq:lb} is no longer a rigid transformation, and contains scale and shear factors, too.

In computer graphics one of the most well-know methods to introduce deformations to models is DQLB. This method blends a surface based on  a set of deformations given in the form of dual-quaternions and weight parameters ~\cite{kavan2007skinning}. Therefore, each joint in the model will be assigned with a dual-quaternion exposing its pose and a wight to define the extent of its effecting region on the model. Given a model with joints $\mathbf{v} = \{v_1, \dots, v_n\}$, dual-quaternions $\mathbf{dq} = \{dq_1,\dots, dq_n \}$ and corresponding weights $\mathbf{w} = \{ w_1, \dots, w_n \}$ we can compute normalized linear combination of all affecting nodes on a specific joint using \eqref{eq:dqbl}.

\begin{equation}
\label{eq:dqbl}
\mathbf{DQLB}(v_i;\mathbf{w};\mathbf{dq}) = \dfrac{w_1 dq_1+ \dots + w_n dq_n}{\lVert w_1 dq_1+ \dots + w_n dq_n \rVert} 
\end{equation}

where, $\mathbf{dq}$ could be either all the nodes or a subset of nearby nodes/vertices to the $i_{th}$ node. Also the extent of the affecting area can be  controlled via weights set $\mathbf{w}$. A detailed description given in literature \cite{kavan2007skinning} and  \cite{ilie2012efficient}. 

%********************************************************************************************

\section{Ray Tracing}
\label{sec:ray_trcing}

\begin{figure}[h]
	\centering
	\includegraphics[width=0.45\linewidth]{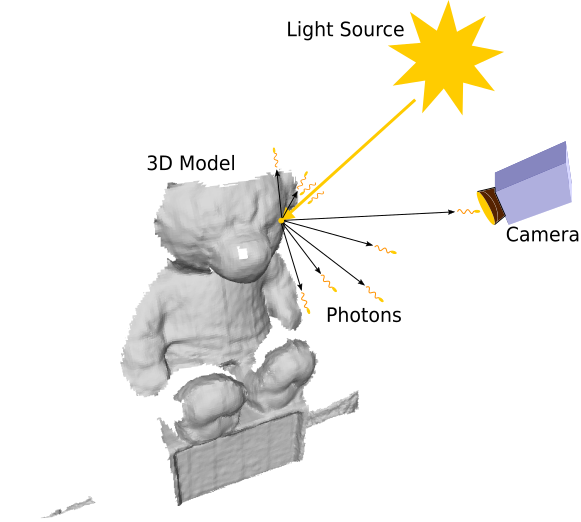}	
	\includegraphics[width=0.45\linewidth]{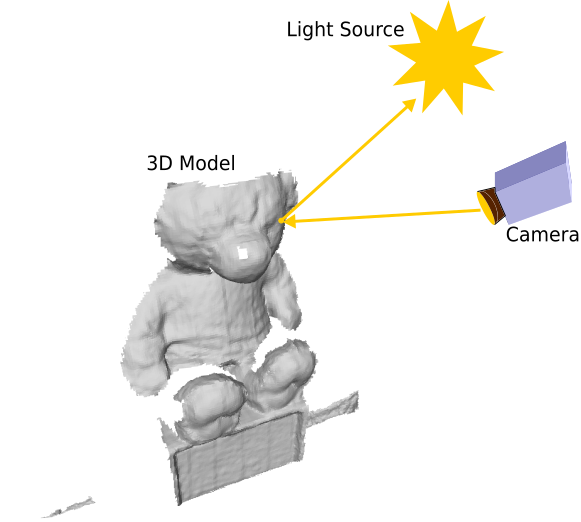}
	\caption[Forward and Backward ray-tracing]{In $left$: a light source is emitting the rays to the environment and via forward Ray-Tracing, algorithm should find out which photon is received by camera to register it on the image pixels, and in $right$: the algorithm only tests few points on the model surface laying along the rays of pixels in the image.}
	\label{fig:frowart_backward_raytracing}
\end{figure}

Ray-tracing is a method to create an image from a given 3D model by projecting back 3D points attributes into image plane of the camera. Considering that only those parts of the object are visible to the camera which receive light from a light source in the environment. There are two major approaches to project photometric and geometric attributes of object into an image. In the first method called forward ray-tracing, we move along the light ray from light source, reflected by the object towards the camera to obtain the color of image pixels. But this is an expensive process, due to the large numbers of light rays that algorithm should check that whether they reach to camera lens or not. On the other-hand , in the second method, we trace rays backwards from the camera to the object surface and then towards the light source, this method is named as backward Ray-Tracing method. Both these methods are shown in \figref{fig:frowart_backward_raytracing}. 

In backward ray-tracing method, rays get emitted from the camera center passing through each pixel. If the emitted ray hits a surface in the 3D environment, we project its light attribute (how much light that specific point receives from light sources in the environment) into the image plane. The hit point can be a shadow point (the point which doesn't receive any light from light sources), a point which is occluded or a normal visible point which reflects the light towards the camera. This process gets repeated for all pixels. Hence, the image resolution will define how expensive would be to whole process of ray-tracing, as long as a constant operation should be repeated for all pixels.

In our implementation, this technique is used in several parts like: generating depth and RGB images from meshes and volumetric representation of the environment.

\begin{figure}[t]
	\centering
	\includegraphics[width=0.8\linewidth]{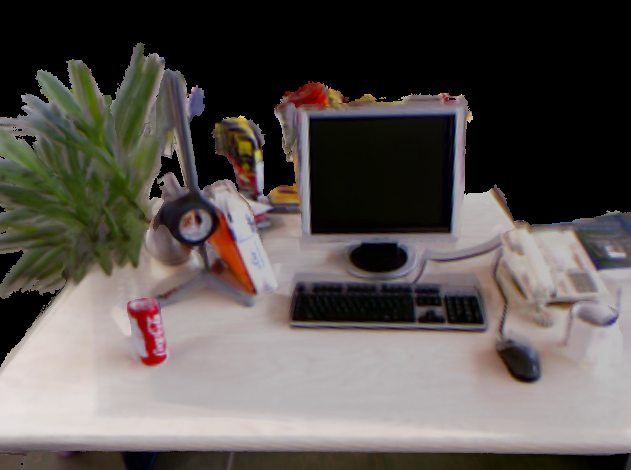}
	\caption[forward and backward ray-tracing results]{A ray-traced RGB image from a given camera pose observing current model status.}
	\label{fig:result_raytracing}
\end{figure}

\clearpage

\subsection{Mesh to Depth Map}
\label{subsec:mesh_to_depth_map}

\begin{figure}[h]
	\centering
	\includegraphics[width=0.8\linewidth]{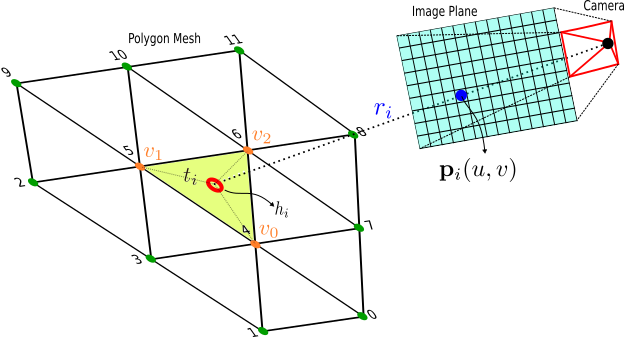}
	\caption[Ray Tracing mesh to depth image]{A camera in a known pose is viewing a polygon mesh where, by finding hit point $h_i$ along pixel ray $r_i$ on the object surface, we can compute the depth value for pixel $\mathbf{p}_i$ in the depth image.}
	\label{fig:mesh_raytracing}
\end{figure}

\begin{figure}[h]
	\centering
	\includegraphics[width=0.48\linewidth]{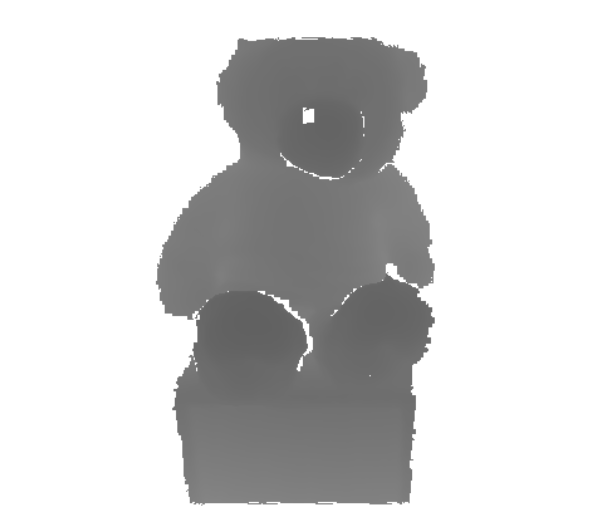}
	\includegraphics[width=0.48\linewidth]{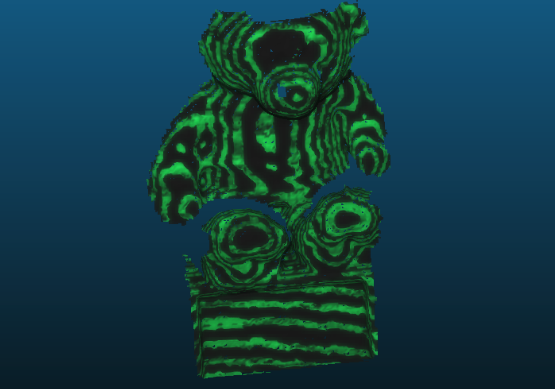}
	\caption[Mesh to depth results]{In $left$: a depth image ray traced from a mesh in a known camera position is shown, and in $right$: a render of target mesh is depicted.}
	\label{fig:mesh_raytracing_result}
\end{figure}

To generate a depth image from a polygon mesh, for each pixel $\mathbf{p}_i$ in the image, we find the triangle $t_i$ which the ray $r_i$ passes though it according to \figref{fig:mesh_raytracing}. Then, based on the position of the hit point $h_i$, we estimate corresponding depth value for each pixel $\mathbf{p}_i$. The position of hit point $h_i$ gets computed using barycentric interpolation \cite{meyer2002generalized} between vertices of triangle $t_i$. An example results of our implementation is shown in \figref{fig:mesh_raytracing_result} where a polygonal mesh is ray traced to a depth image with a known camera position.

%********************************************************************************************
\section{linear Interpolation}
\label{sec:linear_interpolation}
In this work, several data structures such as the nodes in the deformation graph or voxels in a volumetric data representation are discrete functions. Hence, in several instances there is a requirement to generate data at intermediary points as an estimation of the underlying continuous function. A simple way of interpolation is to use the value of the nearest neighbor. However, to have a more accurate estimation, we use linear interpolation techniques. 

Suppose that we have an arbitrary function $\psi(x)$ whose values are known at two points $a$ and $b$. The linear interpolant is the straight line between these points. The function value at the intermediary point $x$ computed using:

\begin{equation}
\label{eq:linear_interpolarion}
\dfrac{\psi(x) - \psi(a)}{x - a} = \dfrac{\psi(b) - \psi(a)}{b - a}
\end{equation}

\begin{figure}[h]
	\centering
	\includegraphics[width=0.5\linewidth]{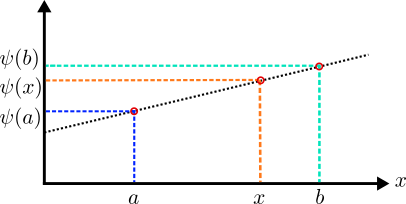}
	\caption[Linear Interpolation in 1-dimensional space]{Linear interpolation in 1-D. The value of function $\psi(x)$ is estimated using two nearby points $a, b$ for which the function values are given. The predicted value is a point on the line between the two known values.}
	\label{fig:linear_interoplation}
\end{figure}

The above equation can be re-written to obtain an expression for the interpolated value $\psi(x)$ (\eqref{eq:linear_w}).
In practice, a weight is given to the function values, proportional to the distance of point $x$ to each side of the line as depicted in \figref{fig:linear_interoplation}.

\begin{equation}
\label{eq:linear_w}
\psi(x) \approx \psi(a) \cdot (b - x) + \psi(b) \cdot (x - a)
\end{equation}

In the 2-D case, the same logic is extended to realize the bilinear interpolation scheme. Here, given the function values at four points $a, b, c$ and $d$,  we look for value of the function $\psi(x,y)$. This is achieved by a series of three 1-D linear interpolations. 

\begin{figure}[h]
	\centering
	\includegraphics[width=0.5\linewidth]{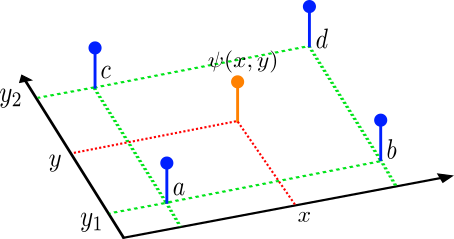}
	\caption[Bilinear Interpolation in 2-dimensional space]{Bilinear Interpolation predicts the value of function $\psi$ at intermediary point $(x,y)$, using three separate linear interpolation.}
	\label{fig:bilinear}
\end{figure}

First, we linearly interpolate between $(a, b)$ and $(c, d)$ in x-direction to obtain $\psi(y_1)$ and $\psi(y_2)$
respectively (\eqref{eq:bilinear_x}). Then we obtain $\psi(x,y)$ by interpolating between $\psi(y_1)$ and $\psi(y_2)$ in the y-direction using \eqref{eq:bilinear_y}. The order of operations can be altered such that the interpolations in the $x$ and $y$ directions are switched.

\begin{equation}
\label{eq:bilinear_x}
\begin{split}
\psi(y_1) \approx \psi(a) \cdot (b - y_1) + \psi(b) \cdot (a - y_1) \\
\psi(y_2) \approx \psi(c) \cdot (b - y_2) + \psi(d) \cdot (a - y_2)
\end{split}
\end{equation}
and,
\begin{equation}
\label{eq:bilinear_y}
\psi(x) \approx \psi(y_1) \cdot (y_2 - x) + \psi(y_2) \cdot (y_1 - x)
\end{equation}

\begin{figure}[h]
	\centering
	\includegraphics[width=0.28\linewidth]{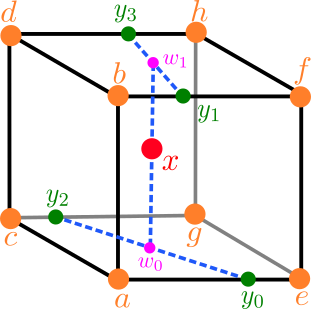}
	\caption[Tri-linear Interpolation in 3-dimensional space]{A cubic space specified with 8 vertices is used to estimate value of function $\psi$ at 3D point $x$ using Tri-linear interpolation including two Bi-linear and one linear interpolation operations.}
	\label{fig:trilinear_cube}
\end{figure}

Similarly, in a 3-D case, where the function values are known at the corners of the cube (\figref{fig:trilinear_cube}), the linear interpolation method can be extended to realize the Tri-linear interpolation.

Let us assign the variables $\hat{x}, \hat{y}$ and $\hat{z}$ to the difference between two points in direction of axes $x, y$ and $z$ respectively as:
 
\begin{equation}
\label{eq:trilinear_distances}
\begin{split}
x_d &= (x - x_0)/(x_1 - x_0) \\
y_d &= (y - y_0)/(y_1 - y_0) \\
z_d &= (z - z_0)/(z_1 - z_0)
\end{split}
\end{equation}

where, $x_0, y_0$ and $z_0$ indicate vertices below intermediary points $x, y$ and $z$ in each axis and $x_1, y_1$ and $z_1$ denote to the upper boundaries respectively \cite{bourke1999interpolation}. Next step would be interpolation along one of the main axes, for example in case of starting in $x$ direction we will have:

\begin{equation}
\label{eq:triInter_y}
\begin{split}
y_0 = a(1 - x_d) + ex_d \\
y_1 = b(1 - x_d) + fx_d \\
y_2 = c(1 - x_d) + gx_d \\
y_3 = d(1 - x_d) + hx_d
\end{split}
\end{equation}

Next, we interpolate in $y$ direction to find $w_0$ and $w_1$ , and the same method is used in $z$ direction to generate interpolated value at point $x$ as follows:

\begin{equation}
\label{eq:triInter_x}
\begin{split}
w_0 = y_0(1 - y_d) + y_2y_d \\
w_1 = y_1(1 - y_d) + y_3y_d \\
\end{split}
\end{equation}

\begin{equation}
\label{eq:triInter_z}
\begin{split}
x = w_0(1 - z_d) + w_1z_d \\
\end{split}
\end{equation}

%********************************************************************************************
\section{K Dimensional Tree (KD-tree)}
\label{sec:k_dimensional_tree}

A KD-tree is a data structure based on a binary search tree frequently used to organize a set of K-dimensional points in a tree shaped structure. It divides the K-dimensional space into partitions which can be accessed through an specific search algorithm. The main usage of this specific data structure is when we need to find the closest or K nearest neighbors of an specific input data in between a given data-set with same data type. For a  large collection of data, the KD tree offers a fast search capability as compared to a naive search. 
In our use-case, we build a KD-tree for the nodes in the  deformation graph (\secref{sec:deformation_graph}). This is necessary
as several queries for obtaining the K nearest neighbors of a node are necessary during the optimization routine.  

In practice, this data structure reorders the data in some layers, where in each layer points are split based on one of their dimensions. For instance, if we start with first dimension of a 3D point which is the $x$ dimension, second layer will be based $y$ and so on. All nodes at the lower layer are called leafs and others are called root nodes From one layer to the next one, the points get rearranged either in the left or right side of root nodes, whether they have bigger or smaller values at that specific dimension. The most efficient way to build a KD-Tree is to use space partitioning method which always picks median points at each level and every other point gets compared w.r.t that point \cite{greenspan2003approximate}.

%********************************************************************************************
\section{Hashing Technique}
\label{sec:hashing_techinque}

In this work, we aim to construct a dense map of the environment. Dense in the sense that, the output will represent the surface and its attributes in fine details (in the order if a few $mm$). Such a representation will require a huge space in the memory to be stored. For example, if the volume in each axis contains 512 voxels, and for each voxel we only store the SDF and weight values, the whole cubic volume will need $512^3 * 16B\approx 1.07 GB$ of RAM to be maintained in the device memory. Even though, in the naive volume management, the majority of the data will carry no information and would denote to either empty or unobserved space rather that target surface. Therefore, utilizing some compression or data-structure is necessary. 
Hence, space partitioning techniques like hashing will increase applications' efficiency while drastically reducing the memory required to store the model.
In this work a hashing function $H(V)$ is used to store and recall TSDF attributes of the voxel $V$ which contains surface information, based on the method proposed in \cite{niessner2013real}. This method is reducing required memory space by scale of 1/8, while the hashing table and complexity of the implementation will be added to the computational complexity too.

%********************************************************************************************
\section{Parallel Computation}
\label{sec:parallel_computation}

\begin{figure}[ht]
	\centering
	\includegraphics[width=0.8\linewidth]{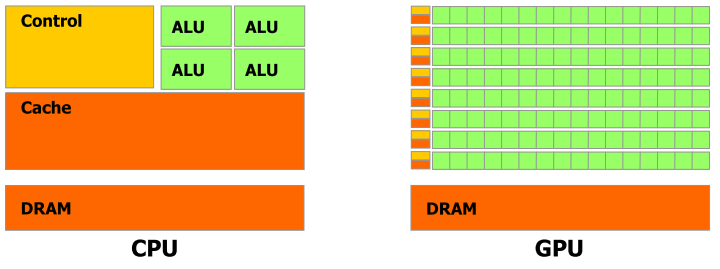}
	\caption[CPU and GPU architecture]{$left$: outlook of CPU architecture with few processing cores and regular global memory structure, $right$: Modern GPUs' architecture, with large number of small processing cores and special shared and global memory structure.}
	\label{fig:cpu_vs_gpu}
\end{figure}

Real-time image processing and mapping techniques require a huge amount of processing resources, due to massive dimension of input data which is meant to be processed. However, most of these operations are independent of each other and can be processed in parallel which makes the overall computation feasible. Nowadays, as multi-core processors and supporting libraries have been advanced a lot, there are different libraries and platforms which can be used to parallelize a program, which most of them use GPU(Graphics Processing Unit) instead of old fashioned CPUs as their main processing unit. In \figref{fig:cpu_vs_gpu} structure of both CPUs and GPUs are shown.

looking at GPU structure, its obvious that it has more processing units than CPU with smaller processing capacity. These small processing units in GPU structure are managed via blocks and grids. Each block contains $n$ processing units defined based on the version of the device, also each grid contains $m$ number of blocks with their-own shared memory. Considering these point, parallel processing is valid for applications with large number of similar operations, while CPUs can perform better in the tasks which contain complex and heavy operations which are highly sequential. In this work, we use CUDA library to parallelize algorithms that have this potential, like Ray-Tracing, extracting mesh, projections and etc.

In CUDA architecture a collection of “streaming multiprocessors” (SM) execute a set of instructions, in parallel on multiple threads on different regions of data managed with array structure. This means, a specific operation which most of the times defines in the main loop, will be executed on a set of selected threads.

For instance, this structure can be utilized in case of having a large array of scalar values (can be an image which is reordered in row-major or column-major shape), which all of its elements need to be processed with similar operation (in case of image can be projection) in parallel. In practice to parallelize our example on an image with resolution of $640 * 480$, we need to have 600 threads per block, this would require us to launch at least 512 blocks to process the entire array in parallel based on $640 * 480 / 512 = 600$. Each of these threads need to know which element of the array to process. Since we have one thread for each element of the array, we use array indices as thread counts too. 

The CUDA run-time defines $threadIdx.x$ to reveal the thread Id within a block and $blockIdx.x$ to define the block Id within the grid. It also exposes $blockDim.x$ to show the dimensions of the block. Putting it all together we can calculate a global id for each thread within the entire grid as following example:

\begin{lstlisting}

__global__void kernel(float *array){ 
	int globalIdx = 
	        blockIdx.x * blockDim.x + threadIdx.x;
	// where N is number of elements in array
	if(globalIdx < N){ 
		array[globalIdx] = array[globalIdx] * 10;
	}
}
\end{lstlisting}

Also, to launch a kernel (functions which execute on GPU), following structure will be used:

\begin{lstlisting}
int threads_per_block = 512;
int thread_blocks =(sensor.rows * sensor.cols + 
                    threads_per_block - 1)
kernel<<<thread_blocks, threads_per_block>>> kernel(Array)
\end{lstlisting}

(For more details, please refer to implementation source in github)

\cleardoublepage{}

\chapter{Dense Rigid SLAM}
\label{chap:dense_rigid_slam}
% !TEX root = ../thesis.tex

The task of creating 3D models of the environment is called mapping. In robotics, this task is often refereed to as the simultaneous localization and mapping (SLAM) problem. A SLAM system can be divided into two major parts, tracking the pose of the camera as it moves (localization) and creating a model of the environment (mapping). The localization part tracks the camera pose at each point in time and based on estimated camera position, mapping part will integrate new scans of the environment into current model.

The localization problem is tackled by registering the current sensor data with the current model of the environment. In our work, we register the RGB and depth images to the current model to estimate the camera pose. Given the pose of the sensor at each point in time, we incrementally build a dense model of surrounding environment based on sensor data perceived over time. 

\begin{figure}[ht]
	\centering
	\includegraphics[width=0.9\columnwidth]{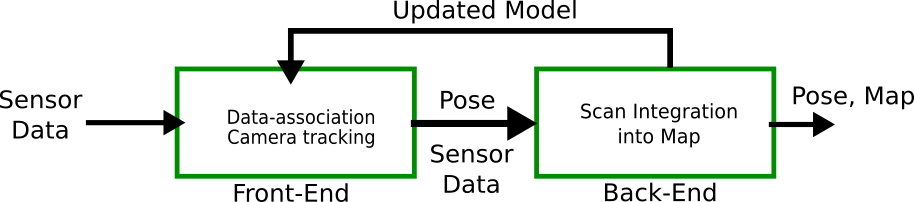}
	\caption[Simultaneous Localization and Mapping(SLAM)]{ The camera pose is estimated by associating the current sensor data to the previously built model. Then, the estimated camera pose used to integrate the data into the model.}
	\label{fig:slam}
\end{figure}

The main pipeline in SLAM is shown in Fig.\ref{fig:slam}, where input data is fed into front-end of the SLAM system where data-association of new scan is performed w.r.t. the current model. Then scans are merged into the current model through given relative transformations obtained by tracking.

In this chapter, we will explain the details of the approach used to track the camera pose and reconstruct the 3D model of environment. 

\section{Related Works}
\label{sec:related_works}
In the context of mapping and geometric reconstruction,  there have been several works including point-cloud based reconstructions using geometric approaches \cite{henry2012rgb} and utilizing photometric methods in \cite{della2018general} offering a framework for multi-cue point cloud registration. Turk \etal , proposed an approach to generate a mesh structure from depth images \cite{turk1994zippered}. The performance of passive and active sensors have been evaluated in \cite{newcombe2010live},\cite{merrell2007real}. Also, the estimation of camera motion known as visual odometry, has received a huge amount of attention \cite{nister2004visual},\cite{davison2007monoslam}. In recent years, availability of commodity RGB-D cameras have led to development of several online mapping systems to perform volumetric map reconstruction \cite{palazzolo2019refusion}, \cite{newcombe2015dynamicfusion}. Such methods have become popular as they could deal with noisy  depth data and obtain detailed reconstructions in an online fashion. 

In our work, we combine both photometric and geometric cues to estimate the camera pose similar to~\cite{izadi2011kinectfusion},~\cite{morency2002stereo}. We do not use any explicit feature detection to build data associations. Instead, we used the idea of projective data association as described in \cite{innmann2016volumedeform}, \cite{stotko2016state}. 

In KinectFusion~\cite{izadi2011kinectfusion}, Izadi \etal proposed a method to incrementally build a volumetric representation of environment  by registering RGB-D frames with a frame-to-model approach. Newcombe \etal, tried to improve the quality of KinectFusion reconstructed model in \cite{newcombe2011kinectfusion}. Also,  Nießner \etal \cite{niessner2013real}, proposed a more efficient version of volumetric scene reconstruction by using voxel hashing method for larger scenes.

In this chapter, we propose an incrementally built volumetric SLAM system, which utilizes a joint geometric and photometric cost function to track the camera pose in the environment via a frame-to-model registration. The volumetric representation integrates new RGB-D frames into the model via hashing techniques to increase memory efficiency. Also, registration pipeline is accelerated via multi-threading techniques to achieve real-time performance.

\clearpage

\section{Camera Pose Estimation}
\label{sec:camera_pose_estimation}
\begin{figure}[h]
	\centering
	\includegraphics[width=0.7\columnwidth]{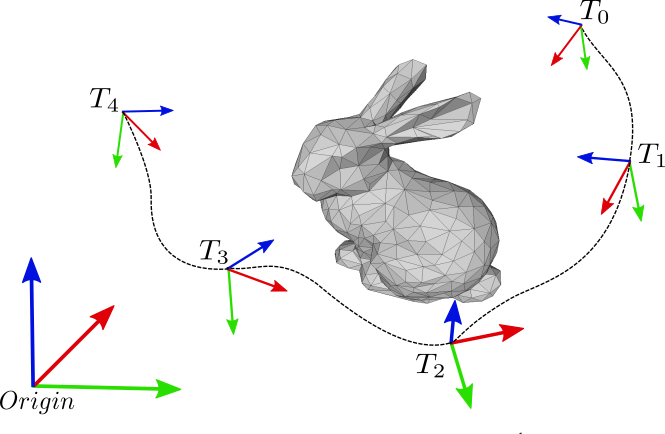}
	\caption[Tracking camera position]{Given two different scans of an object, the registration algorithm can estimate relative alignment parameter between these two frames. By collecting these transformations $\mathbf{T} = \{T_0,\dots,T_n\}$ a 3D model of the target object can be built along with trajectory of the camera in the environment.}
	\label{fig:camera_pose_tracking}
\end{figure}

We estimate 6-DOF pose of the camera shown in \eqref{eq:unknowns-rigid} in 3D world via using rigid-registration techniques.
These techniques find the best match between pairs of scans based on photometric and geometric properties. One case would be using Iterative Closes Point (ICP) to estimate relative alignment parameters when enough geometric information and a proper initial guess along with relatively large overlap is provided. On the other hand, photometric techniques can operate on intensity of regular images to estimate best match via extracting photometric features. By repeating this operation for a series of scans, we can estimate trajectory of the sensor in the environment according to \figref{fig:camera_pose_tracking}

Furthermore, by combining both strategies, faster and more accurate results can be obtained. We define the total optimization function as:

\begin{equation}
\label{eq:erro_total_rigid}
E_{total} = E_{geo} + \lambda E_{pho}
\end{equation}
where, $E_{geo}$ denotes to objective function derived from geometric error minimization and $E_{pho}$ shows the photometric error function. The contribution of each objective function is controlled via weight parameter $\lambda$ that gets a value in range $[0 \sim 1]$.

\begin{figure}[h]
	\centering
	\includegraphics[width=0.9\columnwidth]{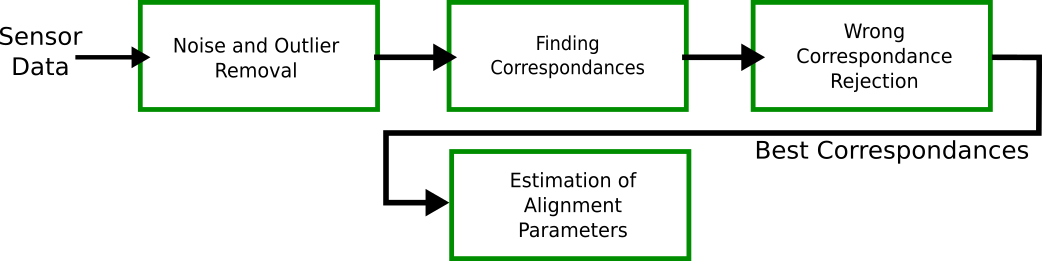}
	\caption[Registration Pipeline]{The new scans before being used to estimate alignment parameters, are de-noised via different filters. Then we find the correspondences between new frames and current model, and filter out inconsistent matches. Finally the optimization pipeline will find the alignment parameter based on correspondences.}
	\label{fig:regis_steps}
\end{figure}

A rigid registration algorithm like ICP iteratively revises the transformation required to minimize error function Eq.\ref{eq:erro_total_rigid}. This algorithm includes different parts which shortly can be named as, noise removal, finding correspondences, rejection of wrong correspondences and finding translation and rotation parameters or the alignment to the current model showing in Fig. \ref{fig:regis_steps}. In this section all mentioned subjects will be covered in details.

This method can be made robust when is combined with robust estimation approaches such as M-estimators \ref{subsec:robust_kernel}.\cite{besl1992method}. In th following we explain each error function as an optimization target and provide enough details about constructing and solving the optimization problem.

\subsection{Geometric-based Error Function}
\label{subsec:geometic_baesd_error_function}
Consider having two RGB-D frames taken at different points in time as frames $\tilde{\mathcal{I}}$ and $\mathcal{I}_{t}$ of dimension $m \times n$ associated with two functions to describe the intensity and depth values at each pixel respectively as $\textbf{I}(p)$ and $\textbf{D}(p)$. In our work, $\tilde{\mathcal{I}}$ is predicted RGB-D frames from current state of canonical model and $\mathcal{I}_{t}$ is new input RGB-D images. Where the parameter $\textbf{p}=[u, v]^\top \in \mathbb{R}^{2 \times mn}$ defines in the pixel domain $\Omega$ of $\mathcal{I}$. Given depth information of each pixel $\mathcal{D}_n \in \mathcal{R}^+$ ,the 3D projection is given using Eq. \ref{eq:projectivecameraS}.

\begin{figure}[h]
	\centering
	\includegraphics[width=0.7\linewidth]{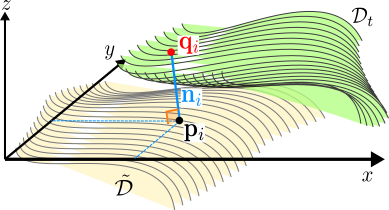}
	\caption[Point-to-Plane error metric]{The distance of point $\mathbf{q}_i$ gets evaluated from point $\mathbf{p}_i$ in  point-set $\mathcal{\tilde{D}}$ along tangent plane normal $\mathbf{n}_i$ at $\mathbf{q}_i$.}
	\label{fig:point_to_plane}
\end{figure}

The optimization function with geometric-based error term, minimizes the sum of the squared error  defined based of Euclidean distances between correspondences in source and target point-sets. This error function can be  defined using different metrics such as point-to-point, point-to-plane and plane-to-plane. Here, we use point-to-plane error function proposed in \cite{chen1991object} that computes distance between two points projected onto the normal direction of the tangent plane associated with one of the points shown in \figref{fig:point_to_plane}.

For instance if, $Q=\{\mathbf{q}_1,...,\mathbf{q}_n\}$ is the source 3D point-set of new input depth map $\mathcal{D}_t$  and $P=\{\mathbf{p}_1,...,\mathbf{p}_n\}$ the corresponding target 3D point-set ray casted to depth image  from current model $\tilde{\mathcal{D}}$ with normals of $N_p=\{\mathbf{n}_1,...,\mathbf{n}_n\}$, the geometric error term with point-to-plane metric can be expressed as: 

\begin{equation}
\label{eq:ICP}
E_{geo} = \sum_{i=1}^{n} \lVert \mathbf{n}_i\cdot(\mathbf{T}_{opt} \cdot \mathbf{q}_i - \mathbf{p}_i)\rVert^2
\end{equation}

where $T_{opt} \in \mathbb{SE}(3)$ is a rigid body transformations:

\begin{equation}
\label{eq:transformation_matrix}
\mathbf{T}_{opt} = 
\begin{pmatrix}
r_{11} & r_{12} & r_{13} & t_x	\\
r_{21} & r_{22} & r_{23} & t_y	\\
r_{31} & r_{32} & r_{33} & t_z	\\
0   & 0   & 0  & 1
\end{pmatrix}
\end{equation}

where, $t$ denotes to linear translation in 3D and $r$ as rotation matrix defines based on  $R(\alpha)$, $R(\beta)$ and $R(\gamma)$ as rotations about axis $x$, $y$ and $z$ respectively as following:

\begin{equation}
\label{eq:rotation}
\mathbf{r}(\alpha, \beta, \gamma) = R_z(\gamma) \cdot R_y(\beta) \cdot R_x(\alpha)
\end{equation}

Hence, by estimating 6 unknowns, $\alpha$, $\beta$, $\gamma$, $t_x$, $t_y$ and $t_z$ the error function \eqref{eq:ICP} can be minimized. Since $\alpha$, $\beta$ and $\gamma$ are arguments of nonlinear trigonometric functions in the rotation matrix $r$, linear least-squares techniques cannot be utilized to minimize the function directly.

To ease the procedure, we can make an assumption that the total rotation on each axis  between constitutive frames are small and  close to zero, then $sin(\theta) = \theta$ and $cos(\theta) = 1$ and transformation matrix can be rewritten as following:

\begin{equation}
\label{eq:transformation_matrix_linear}
T=\mathbf{r}(\alpha, \beta, \gamma),\mathbf{t}(t_x, t_y, t_z) =
\begin{pmatrix}
1 & -\gamma & \beta & t_x	\\
\gamma & 1 & -\alpha & t_y	\\
-\beta & \alpha & 1 & t_z	\\
0   & 0   & 0  & 1
\end{pmatrix}
\end{equation}

Also, using Lie algebra small rigid motions can be represented locally as $\xi \in se3$ Eq.\ref{eq:unknowns-rigid} which is a concise and especial representation in tangent-space of Euclidean group $\mathbb{SE}(3)$. This from of representation can be obtained via \eqref{eq:lie_albegra}, where $\hat{\xi}$ is the corresponding skew-symmetric matrix of $\xi$ in from of Eq.\ref{eq:xi_mat}.

\begin{equation}
\label{eq:unknowns-rigid}
\xi = 
\begin{bmatrix}
\alpha & \beta & \gamma & x & y & z
\end{bmatrix}
\end{equation}

\begin{equation}
\label{eq:lie_albegra}
\mathbf{T}_{opt}(\xi) = exp(\hat{\xi})
\end{equation}

\begin{equation}
\label{eq:xi_mat}
\hat{\xi} =
\begin{bmatrix}
& [\mathbf{r}]_\times & & \mathbf{t}_{3 \times 1} \\
0 & 0 & 0 & 0
\end{bmatrix}  
\end{equation}

where $\textbf{r}$ is a cross-product operator transforming vector $\textbf{r}$ into a $3 \times 3$ skew-symmetric matrix. By substituting \eqref{eq:xi_mat} into Eq.\ref{eq:ICP} the approximation of the objective function would be:

\begin{equation}
\label{eq:objective_func}
E_{geo} \approx \sum_{i=1}^{n} \lVert \mathbf{n}_i \cdot [([\mathbf{r}]_\times  + \mathbf{I}) \cdot \mathbf{q}_i + \mathbf{t} - \mathbf{p}_i]\rVert_2
\end{equation}

and by reorganizing the equation we get:

\begin{equation}
\label{eq:reorg_objective_func}
E_{geo} \approx \sum_{i=1}^{n} \lVert \mathbf{r} \cdot \underbrace{(\mathbf{q}_i \times \mathbf{n}_i)}_{\mathbf{c}_i} + \mathbf{t} \cdot \mathbf{n}_i - \underbrace{(\mathbf{q}_i - \mathbf{p}_i)}_{\mathbf{d}_i} \cdot \mathbf{n}_i \rVert_2
\end{equation}

to minimize the error we need to take the derivative of this equation w.r.t vectors $\textbf{r}, \textbf{t}$ and set the to zero

\begin{equation}
\label{eq:partial_derivative_rt}
\begin{split}
\frac{\partial E_{geo}}{\partial \mathbf{r}} & = \sum_{i=1}^{n} 2 \mathbf{c}_i(\mathbf{r}\cdot \mathbf{c}_i + \mathbf{t} \cdot \mathbf{n}_i - \mathbf{d}_i \cdot \mathbf{n}_i) = 0 \\
\frac{\partial E_{geo}}{\partial \mathbf{t}}  & = \sum_{i=1}^{n} 2 \mathbf{n}_i(\mathbf{r}\cdot \mathbf{c}_i + \mathbf{t} \cdot \mathbf{n}_i - \mathbf{d}_i \cdot \mathbf{n}_i) = 0
\end{split}
\end{equation}

by re-arranging the equations in matrix form and bringing independent parts to the right-side we can get proper form of linear system in form of $\mathbf{A}_{geo}\mathbf{x} = \mathbf{b}_{geo}$, where $\mathbf{x}$ represents unknowns vector and:

\begin{equation}
\begin{split}
\sum_{i=1}^{n} \begin{bmatrix}
\mathbf{c}_i(\mathbf{r} \cdot \mathbf{c}_i) + \mathbf{c}_i(\mathbf{t} \cdot \mathbf{n}_i) \\
\mathbf{n}_i(\mathbf{r}\cdot \mathbf{c}_i) + \mathbf{n}_i(\mathbf{t} \cdot \mathbf{n}_i)
\end{bmatrix}  & =  \sum_{i=1}^{n} \begin{bmatrix}
\mathbf{c}_i(\mathbf{d}_i \cdot \mathbf{n}_i) \\ 
\mathbf{n}_i(\mathbf{d}_i \cdot \mathbf{n}_i)
\end{bmatrix}  \\
\sum_{i=1}^{n} \begin{bmatrix}
\mathbf{c}_i \mathbf{c}_i^\top + \mathbf{c}_i \mathbf{n}_i^\top \\
\mathbf{n}_i\mathbf{c}_i^\top + \mathbf{n}_i \mathbf{n}_i^\top
\end{bmatrix} \begin{bmatrix}
\mathbf{r} \\
\mathbf{t}
\end{bmatrix} & =  \sum_{i=1}^{n} \begin{bmatrix}
\mathbf{c}_i \\ 
\mathbf{n}_i
\end{bmatrix} \begin{bmatrix}
\mathbf{d}_i \cdot \mathbf{n}_i \\
\end{bmatrix}
\end{split}
\end{equation}

resulting target linear equation system which we were looking for

\begin{equation}
\underbrace{
\sum_{i=1}^{\mathbf{n}} 
\begin{bmatrix}
\mathbf{c}_i \\
\mathbf{n}_i
\end{bmatrix} 
\begin{bmatrix}
\mathbf{c}_i^\top  & \mathbf{n}_i^\top\\
\end{bmatrix}}_{\mathbf{A}_{geo}} \mathbf{x}
 =  
\underbrace{
\sum_{i=1}^{\mathbf{n}} 
\begin{bmatrix}
\mathbf{c}_i \\ 
\mathbf{n}_i
\end{bmatrix} \begin{bmatrix}
\mathbf{d}_i \cdot \mathbf{n}_i \\
\end{bmatrix}}_{\mathbf{b}_{geo}}
\end{equation}

As we get to the point which we have the matrices $\textbf{A}_{6 \times 6}, \textbf{b}_{6 \times 1}$ in hand, we  determined unknowns vector $\mathbf{x}$ via robust Cholesky factorization technique \cite{chen2008algorithm}.

\subsection{Photometric Error Function}
\label{subsec:photographic_baesd_error_function}
Camera pose estimation can be addressed via photometric approaches working based on matching pixels intensity values. The main assumption here is that the camera has moved a little bit and the scene is more or less similar to previous frame. Based on this assumption we define photometric error function as \eqref{eq:error_rgb} 

\begin{equation}
\label{eq:error_rgb}
\begin{split}
E_{pho} &= \sum_{i \in \Omega}\lVert
\underbrace{
\mathbf{I}_t(\mathbf{w}(\xi, \mathbf{p}_i)) - \mathbf{I}_{t+1}(\mathbf{p}_i)
}_{\mathbf{e}_{pho}(p_i)}
\rVert^2 \\
E_{pho} &= \sum_{i \in \Omega}\lVert \mathbf{e}_{pho}(\mathbf{p}_i) \rVert^2 
\end{split}
\end{equation}

where, warper function $\mathbf{w}$ warps pixel $(\mathbf{p}_i$ from source image to target image according to \eqref{eq:rgb_warper}. 

\begin{equation}
\label{eq:rgb_warper}
\mathbf{w}(\xi, \mathbf{p}_i) = \pi \left( \mathbf{K} \cdot \mathbf{T}_{vl}(\hat{\xi} ) \cdot \mathbf{P}^{(t+1)}(\mathbf{p}_i) \right)
\end{equation}

To be able to combine results of geometric and photometric error terms in the optimization, we need to define a linear system similar to the one defined for geometric error term. 
First, we linearize error function \eqref{eq:error_rgb}, using Taylor expansion at the current estimate of unknown vector $\xi$ shown in \eqref{eq:rgb_linear}.

\begin{equation}
\label{eq:rgb_linear}
\mathbf{\hat{e}}_{pho}  \approx \mathbf{e}_{pho}^t + \frac{\partial \mathbf{e}_{pho}^t}{\partial \mathbf{\xi}} \cdot \xi + \dots
\end{equation}

where the first term shows the residuals at current estimate of unknown parameters and second part denotes to the derivative of error function \eqref{eq:error_rgb} w.r.t to unknown vector $\xi$ \eqref{eq:unknowns-rigid} resulting in \eqref{eq:rgb_derivative}.

\begin{equation}
\label{eq:rgb_derivative}
\begin{split}
\frac{\partial \mathbf{e}_{pho}}{\partial \mathbf{\xi}}
& = \frac{\partial \mathbf{I}_t}{\partial \mathbf{\pi}} \cdot \frac{\partial \mathbf{\pi}}{\partial \mathbf{K}} \cdot 
\frac{\partial\mathbf{K}}{\partial \mathbf{P}} \cdot 
\frac{\partial \mathbf{P}}{\partial \xi} \\[8pt]
& = \triangledown\mathbf{I}_t \cdot 
\frac{\partial \mathbf{\pi}}{\partial \mathbf{(x, y, z)}} \cdot
\mathbf{K} \cdot 
\frac{\partial \mathbf{\pi}}{\partial \xi}
\end{split}
\end{equation}

the derivative of the intensity function w.r.t to $\mathbf{\pi}$ can be computed as:

\begin{equation}
\label{eq:intensity_derivative}
\triangledown\mathbf{I}_t  = \left(  \frac{\partial \mathbf{I}_t}{\partial \mathbf{x}} , \frac{\partial \mathbf{I}_t}{\partial \mathbf{y}} \right) \in \mathbb{R}^{1 \times 2}
\end{equation}

\begin{equation}
\label{eq:pi_derivative}
\frac{\partial \mathbf{\pi}}{\partial \mathbf{(x, y, z)}} = 
\begin{pmatrix}
\dfrac{1}{z} & 0 & \dfrac{x}{z^2} \\
0 & \dfrac{1}{z} & \dfrac{y}{z^2}
\end{pmatrix} \in \mathbb{R}^{2 \times 3} \\[8pt]
\end{equation}

\begin{equation}
\label{eq:p_derivative}
\frac{\partial \mathbf{P}}{\partial \mathbf{\xi}} = 
\begin{bmatrix}
[\mathbf{P}_i^{(t+1)}]_\times &  & \mathbf{I}
\end{bmatrix} \in \mathbb{R}^{3 \times 6}
\end{equation}

where, the derivation of the equations above is discussed in \cite{gallego2015compact}  and \cite{park2017colored} in more details.
This way we define Jacobian matrices for photometric error term in such a way to agree with geometric part. Finally, we redeclare combined linear systems based on \eqref{eq:erro_total_rigid} which will solve for unknown vector, using geometric and photometric error function as:

\begin{equation}
\label{eq:total_linear_system}
\mathbf{A}_{geo} + \lambda (\mathbf{J}_{pho}^\top \cdot \mathbf{J}_{pho}) = \mathbf{b}_{geo} + \lambda (\mathbf{J}_{pho}^\top \cdot \mathbf{r}_{pho})
\end{equation}

\subsection{Projective Data Association}
\label{subsec:projective_data_association}
\begin{figure}[h]
	\centering
	\includegraphics[width=0.8\linewidth]{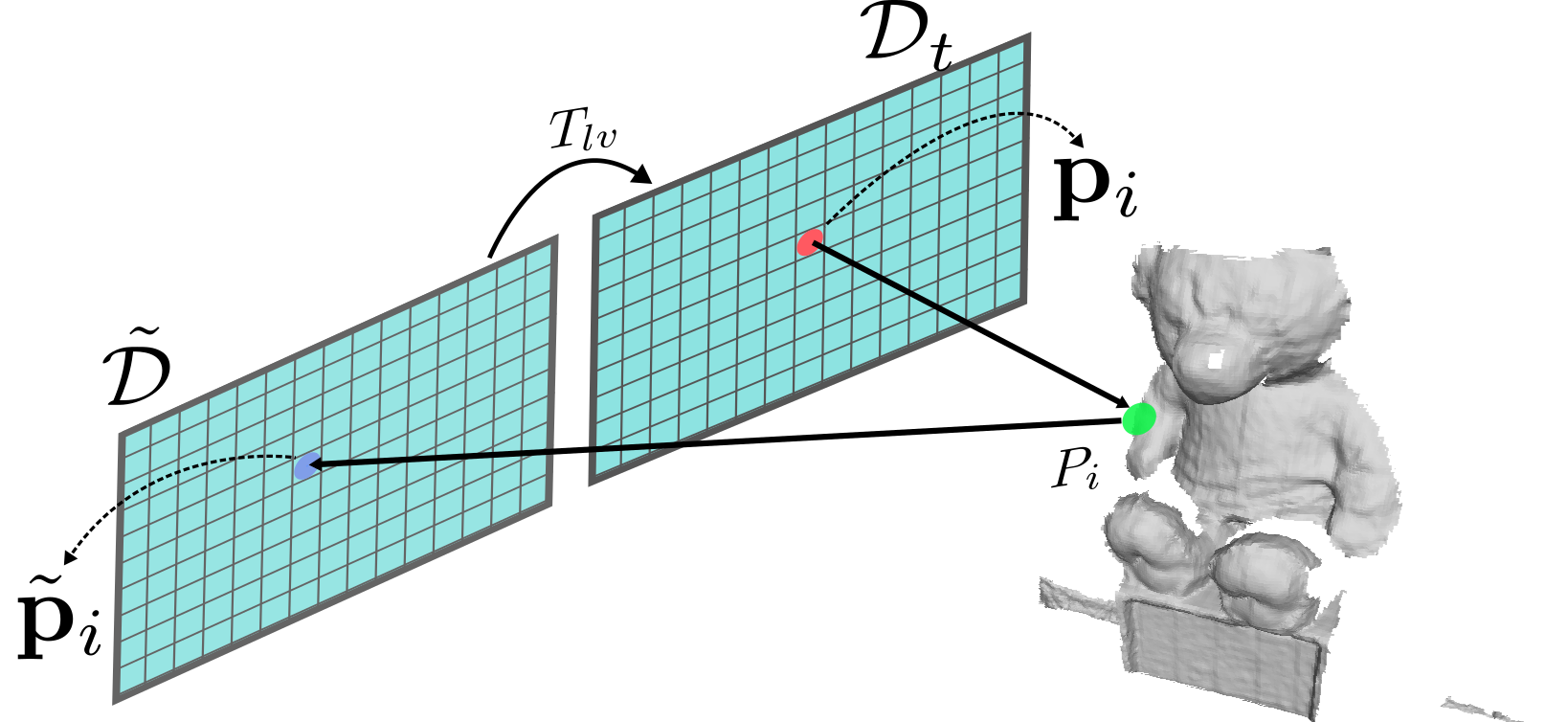}
	\caption[Projective Data Association]{ We project point $\mathbf{p}_i$ from source depth image $\mathcal{D}_t$ to 3D space point $\mathbf{P}_i$ at current camera pose $\mathbf{T}_{vl}$. By back projecting the point $\mathbf{P}_i$ into the target image $\tilde{\mathcal{D}}$ we find the correspondent point $\tilde{\mathbf{p}}_i$.}
	\label{fig:projective_data_assocaition}
\end{figure}

In registration pipeline data-association is a critical task. To select correct correspondences between target an source data-sets different methods like nearest-neighbor, feature-based data association or normal shooting can be used.
In naive approaches algorithm should randomly choose points from one data-set and look for their closest matches in the other point-set, this method is called nearest neighbor (NN) search, which is really expensive to attend for real-time utilization. 
In this work we use Projective Data Association \cite{izadi2011kinectfusion} which is a fast and reliable method in case of having proper initial guess, while it needs to be followed by a strong correspondence rejection step too. Given an estimate of live camera pose $T_{lv}$, we can compute a predicted depth $\tilde{\mathcal{D}}$ map of the current model using Ray-Tracing technique \ref{sec:ray_trcing}. Then the predicted depth map is used as target for input source depth $\mathcal{D}_{t}$ frame which their pixels can be associated together with several steps listed in blow according to  \figref{fig:projective_data_assocaition}.

\begin{enumerate}
	\item Project each pixel $\mathbf{p}_i$ from $\mathcal{D}_t$ to 3D space, generating 3D point $\mathbf{P}_i$. 
	\item Project back 3D point $\mathbf{P}_i$ at latest live camera pose $T_{vl}$ to predicted depth map $\tilde{\mathcal{D}}$ as target image.
	\item Find the index of pixel $\tilde{\mathbf{p}}_i$ where 3d point $\mathbf{P}_i$ is projects to.
	\item Comparing their the normals at source and target images, to reject the inconsistent ones \ref{subsec:surface_normals}.
	\item Building their correspondence pair if normals agree with each other and the distance of their projection is less that a threshold in range of $\pm10cm$.
\end{enumerate}

\begin{figure}[h]
	\centering
	\includegraphics[width=0.6\linewidth,]{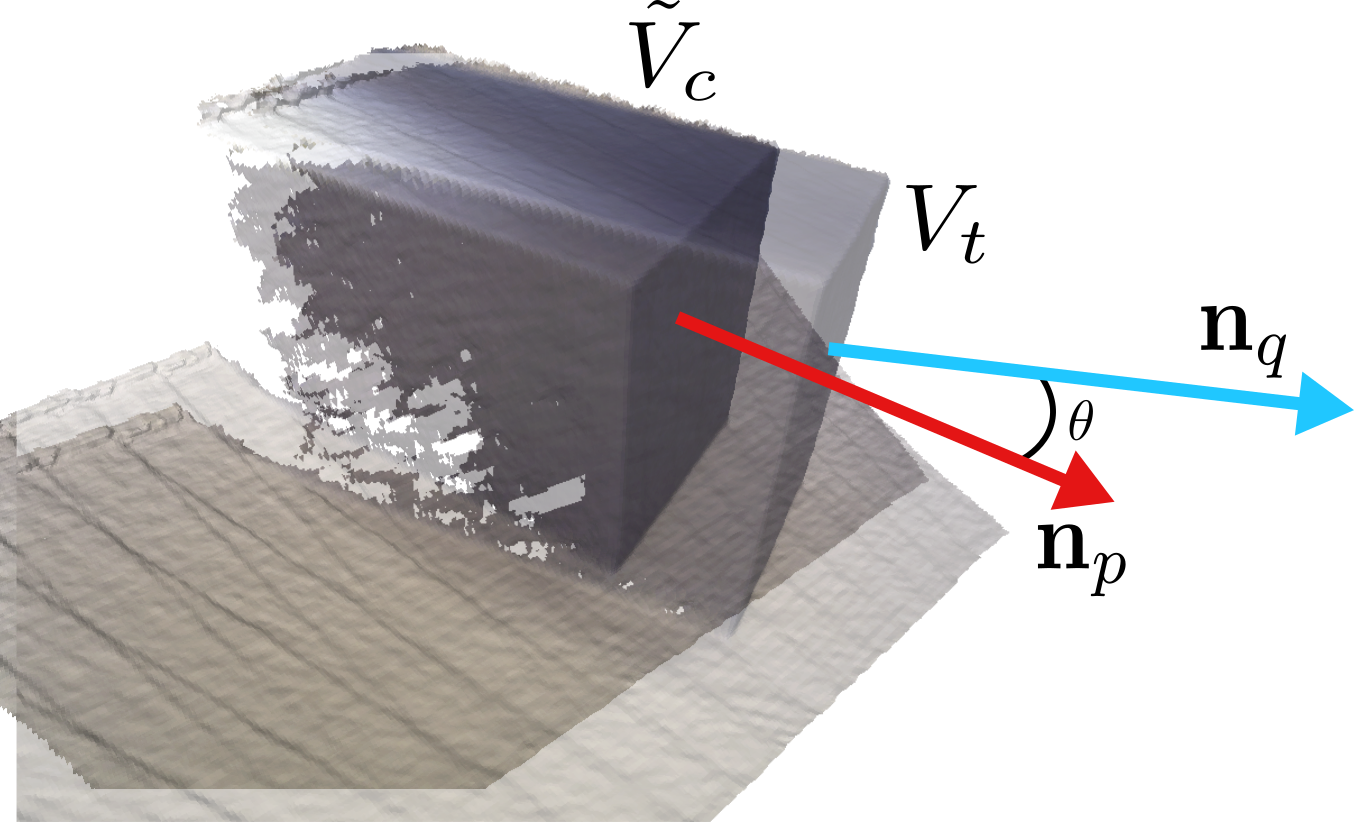}
	\caption[Surface normal evaluation]{The normals at corespondent points in source and target depth maps are compared based on the angle between them to reject the inconsistent pairs of corespondents.}
	\label{fig:normal_rejection}
\end{figure}

In the correspondence rejection step the normals at source and target vertices will be evaluated by their difference in angle using \eqref {eq:normal_rejection}. Where, the pairs with angle difference bigger than $\pm$20\textdegree  will be rejected according to  \ref{fig:normal_rejection}. 

\begin{equation}
\label{eq:normal_rejection}
\theta = \arccos(\mathbf{n}_p \cdot \mathbf{n}_q)
\end{equation}

\subsection{Coarse-to-fine processing}
\label{subsec:coarse_to_fine_processing}
The registration of to RGB-D scans can have a non-convex shape in many situations, for instance if the  initial guess is not proper, which causes the optimization process to get trapped in a local minima. To avoid such situations we used cores-to-fine approach, which tries to solve approximations of the problem with less details and increase the accuracy as it gets closer to global minimum \cite{thevenaz1998pyramid}. As section \ref{subsec:image_pyramid} exploits we down sample input images with pyramid technique. 

Therefore, in subject of optimization objective, problem in higher levels of pyramid are smoother and leads the gradient descent toward true minimum, hence there is less chance to get stuck in a lock minima. Then result of each coarse level is  used to initialize the start point of next level of optimization until the process reaches to the based of pyramid and with only a few iterations can find the global minimum.

\clearpage

\section{Mapping}
\label{sec:mapping}

Our pipeline of mapping integrates new scans into the model based on estimated relative alignment parameters of each new scan to the current model. In context of mapping, we maintain an implicit representation of the environment or object based on input RGB-D scans. Implicit, in the sense that it describes the space around surface based on chosen attributes (distance from nearest surface, color, etc), leaving it up to us to infer surface positions and orientations indirectly \cite{canelhas2017truncated}. Resulting model can be visualized via direct methods like ray casting to images or different 3D object representation methods like triangular mesh \cite{newcombe2012dense}. 

In this section, we will discuss in details how and in which steps we construct the surface representation in our implementation.

\subsection{Signed Distance Field}
\label{subsec:signed_distance_field}

\begin{figure}[h]
	\centering
	\includegraphics[width=0.6\linewidth]{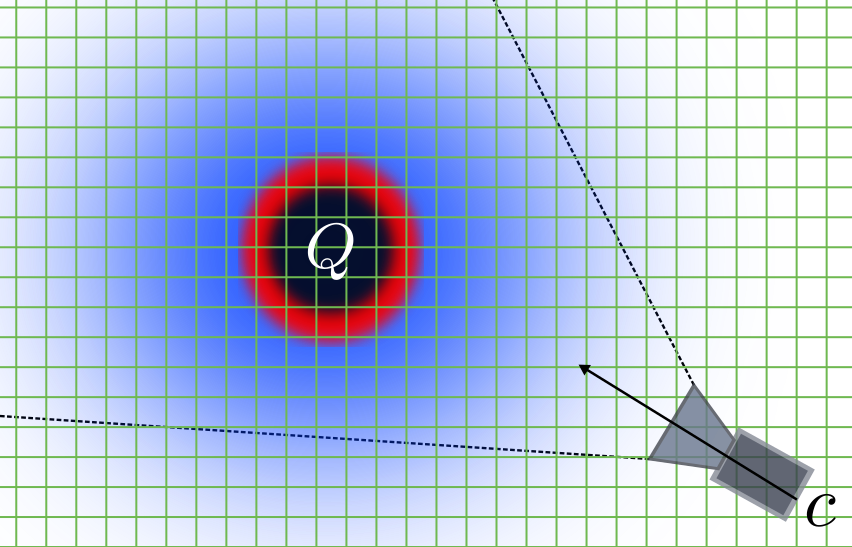}
	\caption[Distance filed]{A distance field is constructed around object $Q$ via depth images provided at point $c$.}
	\label{fig:distnace_field}
\end{figure}

One of the well-studied implicit field functions is the distance field which is used to reconstruct 3D models of objects based on range scans. A field representation can be defined as a space that preserves discrete scalar values in voxelized structures. In case of distance field, a value at any given voxel is interpreted to the distance of that point to the nearest surface. For instance, in  \figref{fig:distnace_field}, a camera at point $c$ is scanning a surface of object $Q$, where the different scalar values of the distance field at each point around the surface is shown with color spectrum. 

A empty grid cell gets a discrete distance value based on its distance to the nearest surface. The nearest surface location is defined based on the range measurements. The distance field in definition is a continuous function, but in practice a sample-based approximation is implemented based on discrete distance function. After constructing the distance field, interpolation techniques \secref{sec:linear_interpolation} are used to retrieve continues estimate of the field.

\begin{figure}[h]
	\centering
	\includegraphics[width=1.0\linewidth]{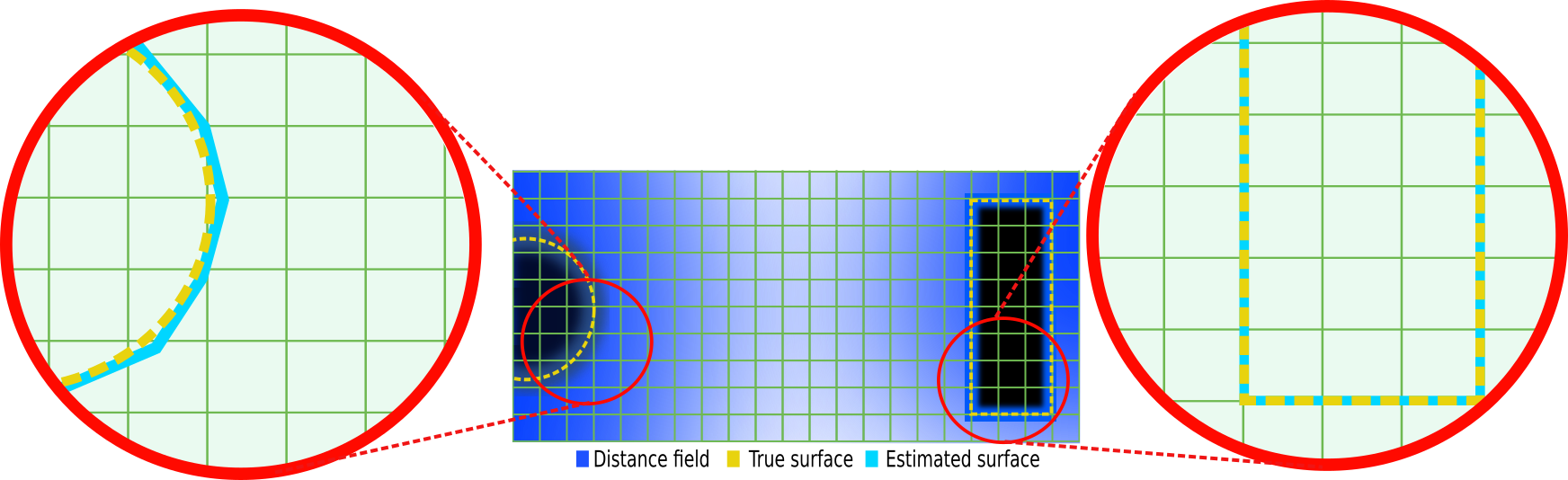}
	\caption[Distance filed problem]{The distance field exploiting two circular and rectangular objects. The grid cells at the broader of the surface are causing artifacts in true surface representation by degrading the accuracy of surface position in each cell. In case of rectangular shapes grid cell provide more precise representation in comparison with shapes with curvature.}
	\label{fig:df_problem}
\end{figure}

The naive distance field, only stores distance of each pixel/grid cell to the scanned surface. This can cause ambiguity in surface position an normal extraction. For instance if, the geometry is not completely aligned with the grid cells, the surface should be approximated and this causes losings precision, shown in \figref{fig:df_problem}. 

This type of field representation can have acceptable performance a long as objects are not completely scanned. In case of having $360^\circ $ scans of an object, due to the conflict between values of different scans, the field will lose the accuracy. This failure is because the surface scanned at each point will cause an increase in grid cells values and gradually extracting the surface will be impossible also with same scalar values at both sides of a surface, extracting true direction of it would be ambiguous. To deal with this problem, the signed distance fields has been introduced. This method signs all cells in the grid upon their location, whether they are located in front (positive) or back of the surface (negative). This way to extract the surface algorithm only needs to find the point where the sign of distance function changes (zero-crossing point). Obtaining a signed distance field is expensive operation in the scene that the method assumes the field expands to infinity from each direction. Hence, Curless \etal~\cite{curless1996volumetric}, proposed a method called projective signed distance function which defines distance field around a surface within a range called truncation area only along the direction of measurements rays (line-of-sight).

\subsection{Truncated Singed Distance Function}
\label{subsec:truncate_singed_distance_function}
\begin{figure}[h]
	\centering
	\includegraphics[width=0.8\linewidth]{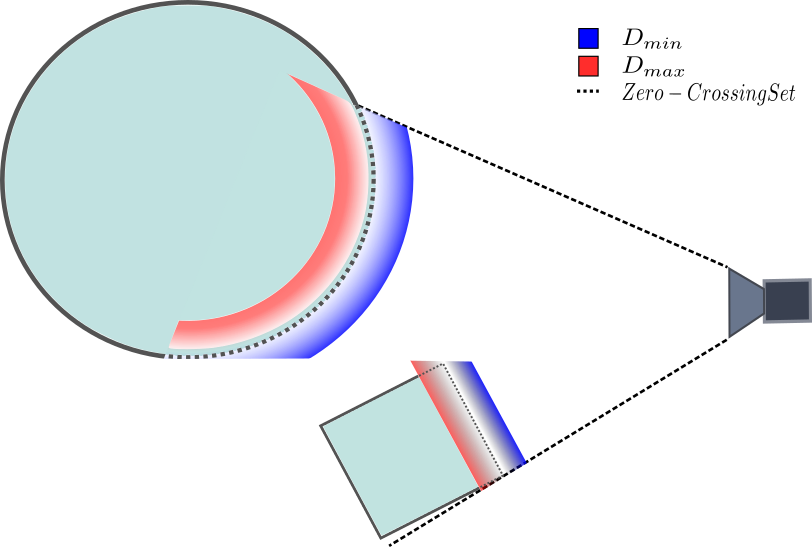}
	\caption[Truncate signed distance function]{A camera is observing two object in its field of view and forming TSDF around scanned true surface withing truncation area denoted with blue (outside of object and positive SDF value) and red (inside the object and negative SDF values) colors.}
	\label{fig:tsdf}
\end{figure}

This method integrates depth images into a volume, via partially updating the signed distance field along line-of-sight direction around the true surface depicted in \figref{fig:tsdf}. This method integrates new values only in a truncated region between $D_{min}$ and $D_{max}$ and near the current estimate of the surface instead of having full range field around the observed geometry ~\cite{curless1996volumetric}. 

To update a specific grid cell, they propose a weighted signed distance function, where, each point in the space is described with a SDF value estimated from a continuous implicit distance function $\mathbf{D(x)}$, and a wight function $\mathnormal{W(x)}$. Using these functions, the SDF value at each point $x$ calculates based on a weighted combination of points laying on the line-of-sight, passing through the point $x$ via \eqref{eq:sdf_d} and \ref{eq:sdf_w}. Therefore, on a discrete voxelized grid, the corresponding iso-surface can be extracted on $TSDF(x) = 0$. 

\begin{equation}
\label{eq:sdf_d}
\mathnormal{TSDF_{i+1}(x)} = \dfrac{\mathnormal{W_{i}(x)TSDF_{i}(x) + w_{i+1}(x) d_{i+1}(x)}}{\mathnormal{W_{i}(x) + w_{i+1}(x)}}
\end{equation}

\begin{equation}
\label{eq:sdf_w}
\mathnormal{W_{i+1}(x)} =\min(\mathnormal{W_{i}(x) + w_{i+1}(x)}, W_{max})
\end{equation}
where, $d_i$ and $w_i$ are the signed distance and weight functions from the $i_{th}$ range image and $W_{max}$ represents maximum weight limit. 

\begin{figure}[h]
	\centering
	\includegraphics[width=0.7\linewidth]{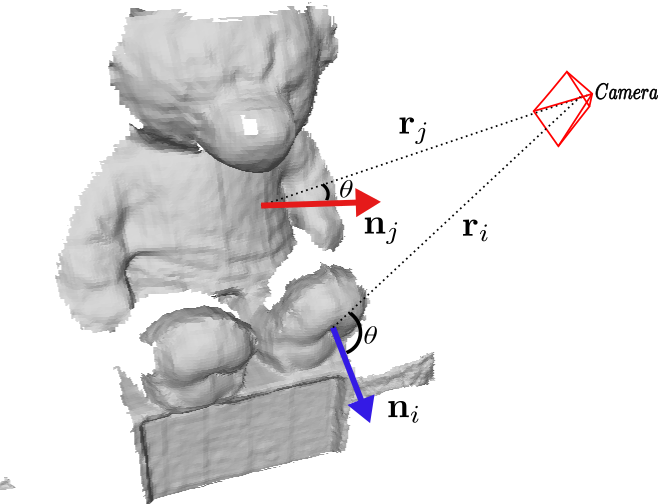}
	\caption[Surface Normals at Volumetric integration]{The normals $\mathbf{n}_i$ and $\mathbf{n}_j$ are evaluated based on their angle w.r.t the rays $\mathbf{r}_i$ and $\mathbf{r}_j$ to assign proper weights.}
	\label{fig:tsdf_weight_normal}
\end{figure}

Assigning weights are necessary to distinguish how reliable SDF values are at each point. In practice, range of assigned wights should be set relative to sensor type. In case of using optical sensors, the weight depends on the dot product between  each vertex normal and the line-of-sight according to \figref{fig:tsdf_weight_normal}. Hence, uncertainty is greater when the $\theta$ is bigger. Furthermore, keeping track of the weights helps to robustness to the outliers as the signal to noise ratio increases over the time by integration more scans into the volume which will result finer appearance of details\cite{curless1996volumetric}.

The corresponding color of RGB pixel is also projected along the line-of-sight and gets integrated into the volume using \eqref{eq:tsdf_rgb} and gets stored as an attribute for each voxel. Thereby, this extra information enables us to texture the mesh and use of intensity information during the registration. 

\begin{equation}
\label{eq:tsdf_rgb}
\mathnormal{\mathbf{I}_{i+1}^{rgb}(x)} = \dfrac{\mathnormal{W_{i}(x)\mathbf{I}_{i}^{rgb}(x) + w_{i+1}(x) i_{i+1}^{rgb}(x)}}{\mathnormal{W_{i}(x) + w_{i+1}(x)}}
\end{equation}

where, $\mathbf{I}^{rgb}$ and $i^{rgb}$ denote to the individual channels of color attributes in volume and RGB frames, respectively.

Let us assume a camera pose is know and new frames of RGB-D sensor are available. The question is that, how new depth and RGB frames should be integrated into the existing volume. The following steps should be taken to integrate a specific pixel of depth and RGB images into the current estimate of volumetric model.

\begin{enumerate}
	\item Finding the line-of-sight of each pixel relative to current camera position.
	\item Finding correspondent position of scanned surface in 3D space of the volume, using depth values and line-of-sight of each individual pixel.
	\item Traversing through $D_{min}$ to $D_{max}$ (in truncation region) and updating voxel attributes using functions $TSDF(x), W(x)$ and $\mathbf{I}^{rgb}(x)$.
\end{enumerate}

\section{Surface Extraction and Visualization}
\label{sec:surface_extraction_and_visualization}

we need to extract a 3D models from the implicit representation. There are two approaches, direct ray-casting or surface extraction using polygon-based methods.
The direct method, gives an estimate of the volume content in latest estimated camera position. This estimate is an image, which can depict different attributes like surface normals, predicted depth or texture containing color information. To address this task, different methods can be employed like rasterization \cite{akenine2018real} and ray tracing \ref{sec:ray_trcing}.

Whereas sometimes, constructing a polygon-based representation of the surface would be more convenient, in order to be used in processes like applying deformations \cite{kavan2007skinning} and non-rigid optimizations.
 
\subsection{Polygon Mesh Extraction}
\label{subsec:polygon_mesh_extraction}
\begin{figure}[h]
	\centering
	\includegraphics[width=0.5\linewidth]{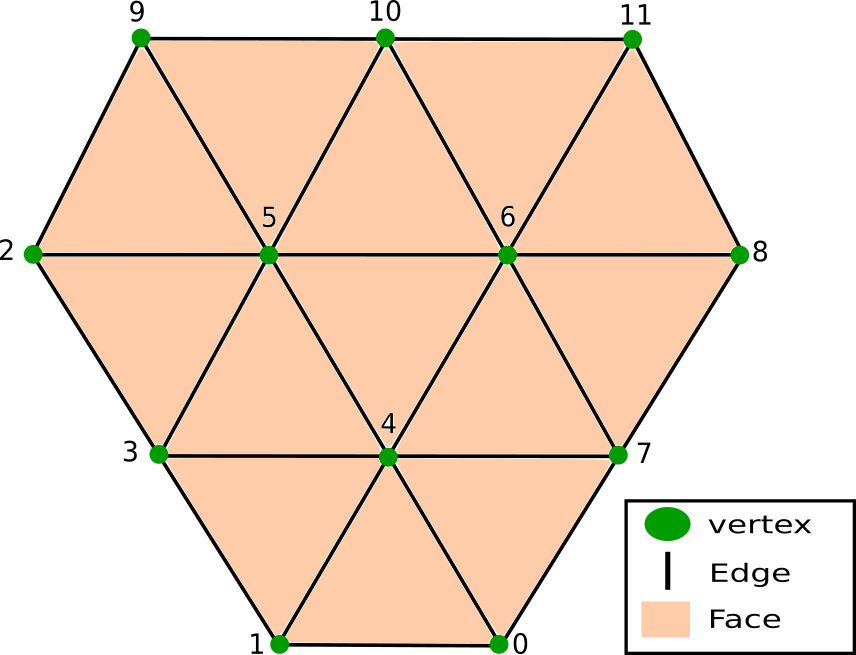}
	\caption[Simple mesh]{A simple mesh with 12 vertices and 13 faces (triangles) is shown.}
	\label{fig:mesh_simple}
\end{figure}

The meshes or generally polygon meshes are collection of vertices, edges and faces that make up a 3D object and could contain texture information too. The \figref{fig:mesh_simple} shows simple construction of faces based on vertices and lines(edges). 

Typically faces consist of triangles (triangle mesh), quadrilaterals (quads), or other simple convex polygons. There are several techniques to generate polygon meshes which most of them built up on the principle of Delaunay Triangulation \cite{edelsbrunner2001geometry}. In practice generating mesh involves subdivision of a continuous geometric space into discrete geometric and topological cells, where each cell models or approximates a specific part of the given surface \cite{frey2007mesh}.

\subsubsection{Marching Cubes}
\label{subsubsec:marching_cubes}

\begin{figure}[h]
	\centering
	\includegraphics[width=0.7\linewidth]{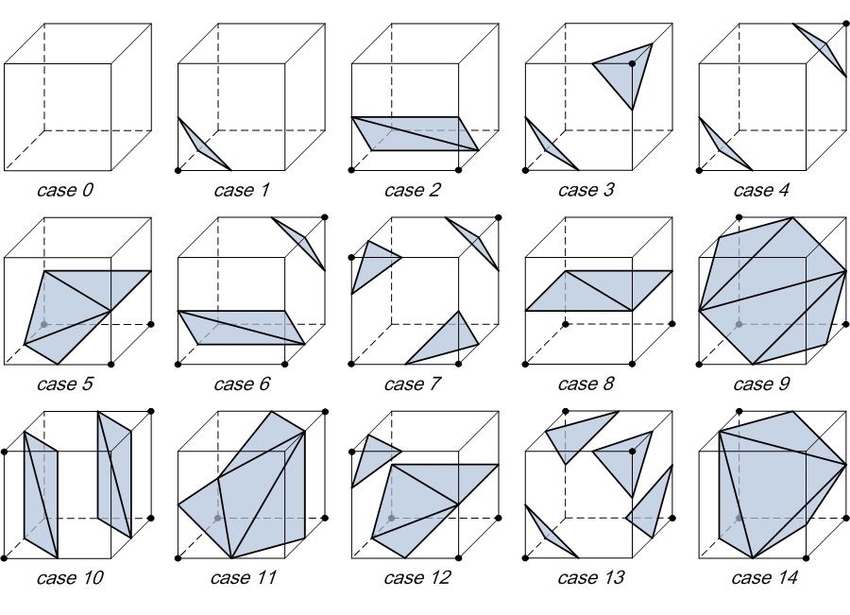}
	\caption[Polygon configurations in a cube]{Several combination of surface interventions in cubes are shown. A cube can be completely inside or out side of the object which in these cases cube would not contain any triangle. Also, a cube can contain several faces at the same time based on edges which are intersected via zero-crossing level set \cite{long2015marching}.}
	\label{fig:mesh_table}
\end{figure}

Marching cubes can be names as the most popular algorithm to generate triangular meshes, due to its simplicity. This method first estimates the bounding box around the object, then divides the space into an arbitrary number of cubes using 8 neighbor voxel centers in the volume. The next step involves testing each edge/side of the cube to find the intersection with object boundaries (zero-crossing level). A cube can be completely inside the object or partially intersecting with the object surface or it can be completely outside of it. In case which object surface is intersecting with cube, algorithm approximates surface in that specific position using a pre-computed intersection table. There are only a few possible configurations for surface intervention in a cube, which some of them shown in \figref{fig:mesh_table}

\begin{figure}[h]
	\centering
	\includegraphics[width=0.8\linewidth]{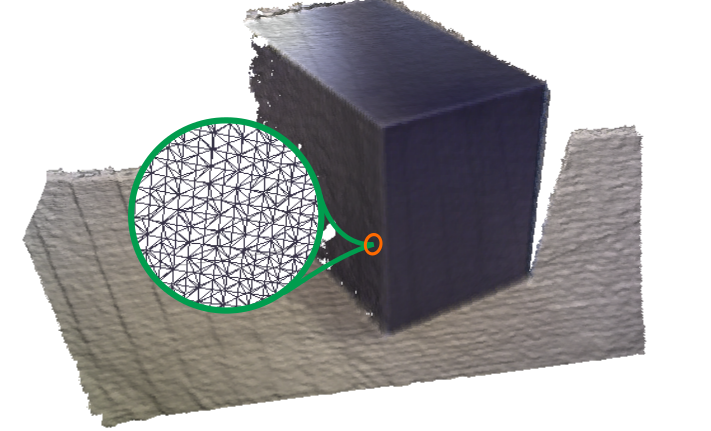}
	\caption[Marching Cubes output]{An output of marching cubes is shown where in zoomed circle generated triangles are visible.}
	\label{fig:exported_mesh}
\end{figure}

In our work, we extract mesh from a three-dimensional volumetric space containing iso-surface information, which is discussed in details in \secref{subsec:truncate_singed_distance_function}. The volume stores iso-surface values in a discrete voxelized space. The marching cube algorithm forms cubes using center of 8 neighbor voxels and finds the best approximation of the surface in that cube. In practice finding closest approximation to the surface is done by creating a pre-calculated array of 256 possible polygon configurations ($2^8=256$). The index of intersection type in the configuration array gets computed based on a 8-bit number, where each voxels' scalar value can alter state of one bit. If the scalar's value is higher than the iso-surface value it will be considered as inside the surface, then the appropriate bit is sets to one, while if it is lower (outside), it is sets to zero. The final value, after all eight scalars are checked, is the actual index in the polygon array \cite{newman2006survey}. An example output is shown in \figref{fig:exported_mesh}.

\section{Results and Evaluation}
\label{sec:results_and_Evaluation}

The main contribution of this chapter is a dense mapping pipeline which is able to construct a map of the environment based on geometric and photometric information. The evaluation is based on estimating root-mean-square (RMS) error  using \eqref{eq:rmse}, as well as considering relative pose error (RPE) that measures local accuracy of generated trajectory in a fixed time interval $\Delta$ using \eqref{eq:rpe} \cite{sturm2012benchmark}.   

\begin{equation}
\label{eq:rmse} 
E_{rms} = \sqrt{\dfrac{\sum\limits_{i=1}^n (\mathbf{trans}(\hat{\mathbf{T}}_i) - \mathbf{trans}(\mathbf{T}_i))^2}{n}}
\end{equation}

\begin{equation}
\label{eq:rpe} 
E_{rpe} = (\hat{\mathbf{T}}_i^{-1} \cdot \hat{\mathbf{T}}_{i + \Delta})^{-1}(\mathbf{T}_i^{-1} \cdot \mathbf{T}_{i + \Delta})
\end{equation}

where $n$ show the number of frames and $\hat{\mathbf{T}}_i$ and $\mathbf{T}_i$ denote to predicted camera pose and reference camera pose respectively for $i_{th}$ frame. Also $\mathbf{trans}(T)$ gives the translation part of a transformation matrix.

The experiments shows the capability of the our approach to deal with noises in the dataset. The Depth frames can contain our-of-range and zero range values, which both can disturb the results of the fusion. To reduce negative effect of such noises we ignore pixels with zero and out-of-range values in the process of camera tracking and volumetric fusion, also using bilateral filter we smooth input depth measurements and fill small holes in depth frames.  

We have tested our approach on static scenes from TUM collection \cite{sturm2012benchmark} which their result are shown in \tabref{tab:tum_results}.

\begin{table}[h!]
	\begin{center}
		\caption{Rigid dense SLAM results on TUM datasets.}
		\label{tab:tum_results}
		\begin{tabular}{l|c|c} % <-- Alignments: 1st column left, 2nd middle and 3rd right, with vertical lines in between
			\textbf{Name} & \textbf{RMSE} & \textbf{RPE} \\
			 & $ (m) $ & $ (m) $ \\
			\hline
			TUM $fr2-xyz$ & 0.1274 & 0.1482 \\
			TUM $fr2-rpy$ & 0.0509 & 0.0823 \\
			TUM $fr2-loh$ & 0.1124 & 0.1224 \\
			TUM $fr2-cabinet$ & 0.0686 & 0.0927 \\
			TUM $fr2-teddy$ & 0.1090 & 0.1207 \\
		\end{tabular}
	\end{center}
\end{table}

We also tested our implementation on virtual dataset from NUIM which result are given in \tabref{tab:nuim_results}.

\begin{table}[h!]
	\begin{center}
		\caption{Rigid dense SLAM results on NUIM datasets.}
		\label{tab:nuim_results}
		\begin{tabular}{l|c|c} % <-- Alignments: 1st column left, 2nd middle and 3rd right, with vertical lines in between
			\textbf{Name} & \textbf{RMSE} & \textbf{RPE}\\
			 & $ (m) $ & $ (m) $ \\
			\hline
			IVR $taj1$ & 0.08629 & 0.1148 \\
			IVR $taj2$ & 0.06962 & 0.0975 \\
		\end{tabular}
	\end{center}
\end{table}

\begin{figure}[h]
	\centering
	\includegraphics[width=0.42\linewidth]{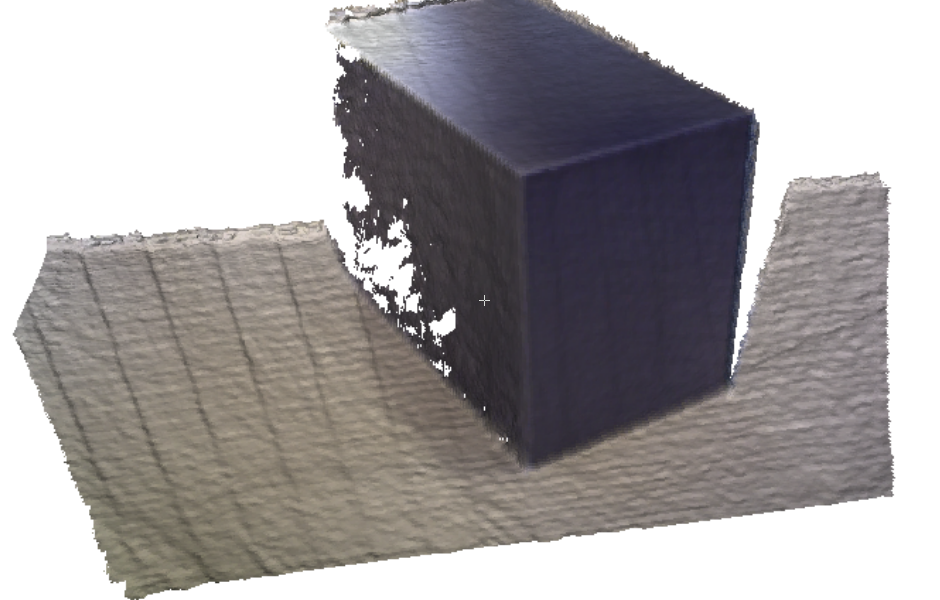}
	\includegraphics[width=0.42\linewidth]{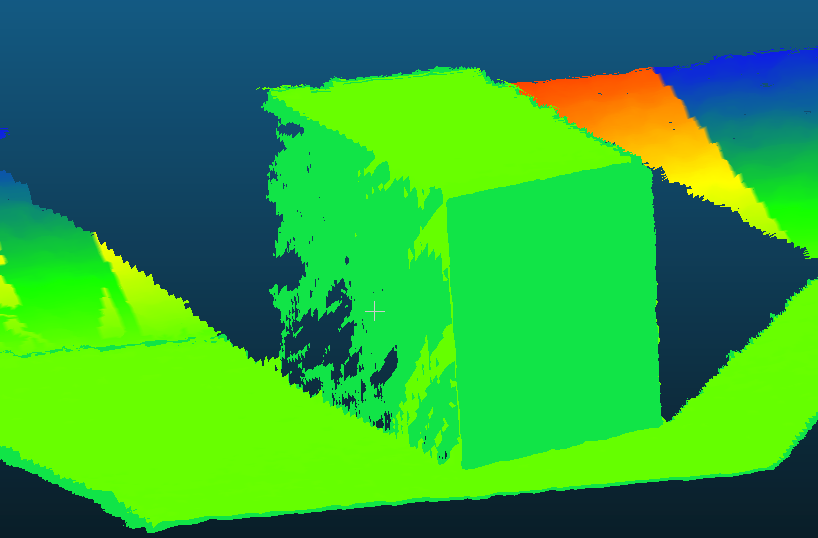}
	\includegraphics[width=0.07\linewidth]{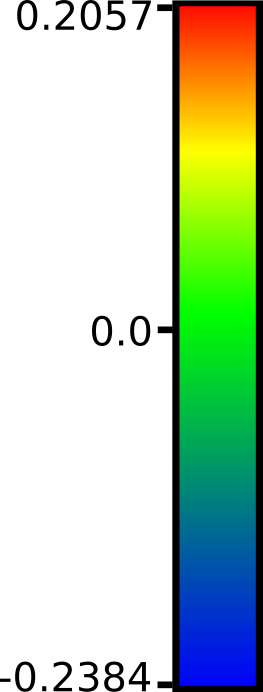}
	\caption[Reconstruction of $fr3-cabinet$ RGB-D data-set]{$left$: 3D reconstruction of $fr3-cabinet$ data-set from TUM collection, $right$: surface evaluation w.r.t reference model coded in color spectrum. The camera track is a circular trajectory with various types of motions, but due to simplicity of the observed geometry, the reconstruction generates acceptable results.}
	\label{fig:res_cube}
\end{figure}
\begin{figure}[h]
	\centering
	\includegraphics[width=0.42\linewidth]{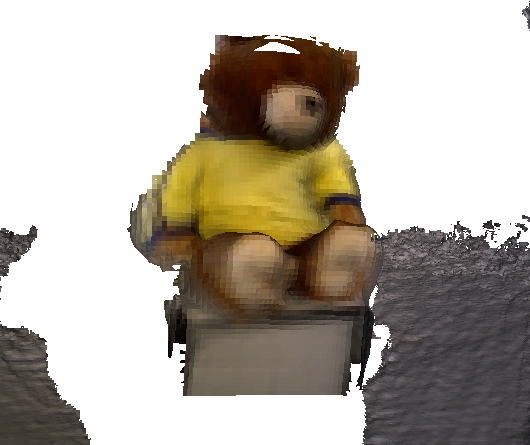}
	\includegraphics[width=0.42\linewidth]{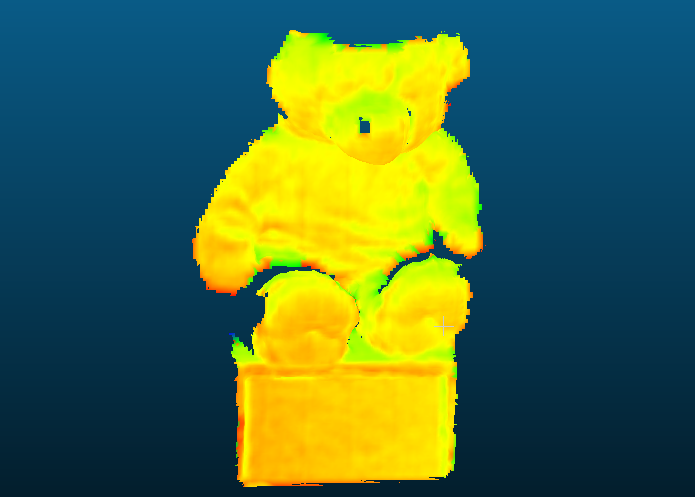}
	\includegraphics[width=0.07\linewidth]{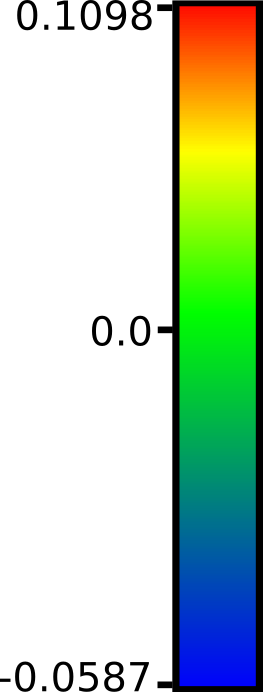}
	\caption[Reconstruction of $fr3-teddy$ RGB-D data-set]{In $left$: RGB reconstruction of teddy bear from $fr3-teddy$ data-set is shown.This data-set contains sharp motions and a lot of vibrations which cause the tracking loss the accuracy.}
	\label{fig:res_teddy}
\end{figure}
\begin{figure}[h]
	\centering
	\includegraphics[width=0.42\linewidth]{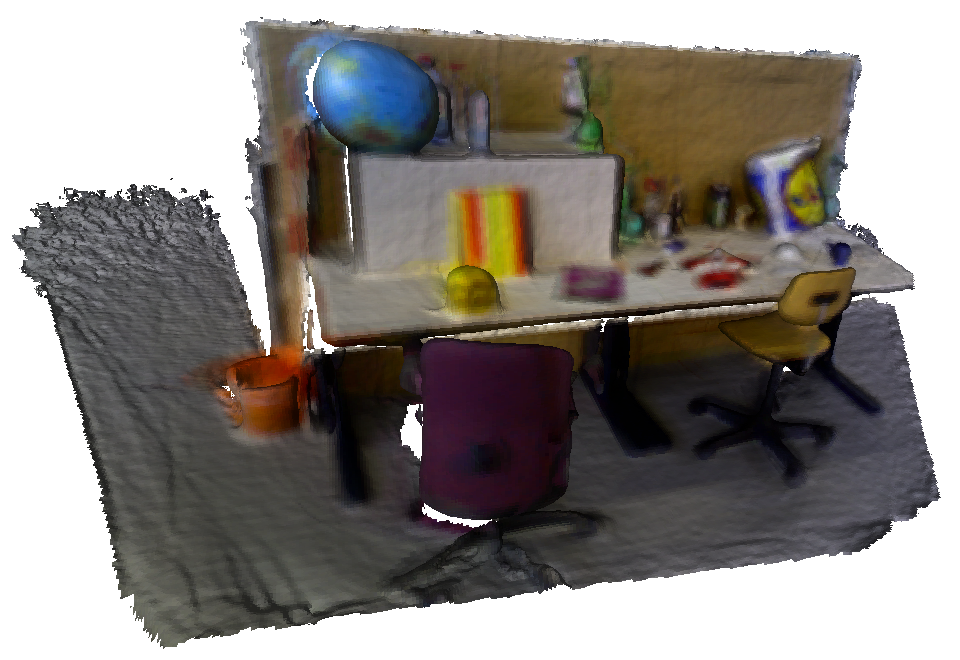}
	\includegraphics[width=0.42\linewidth]{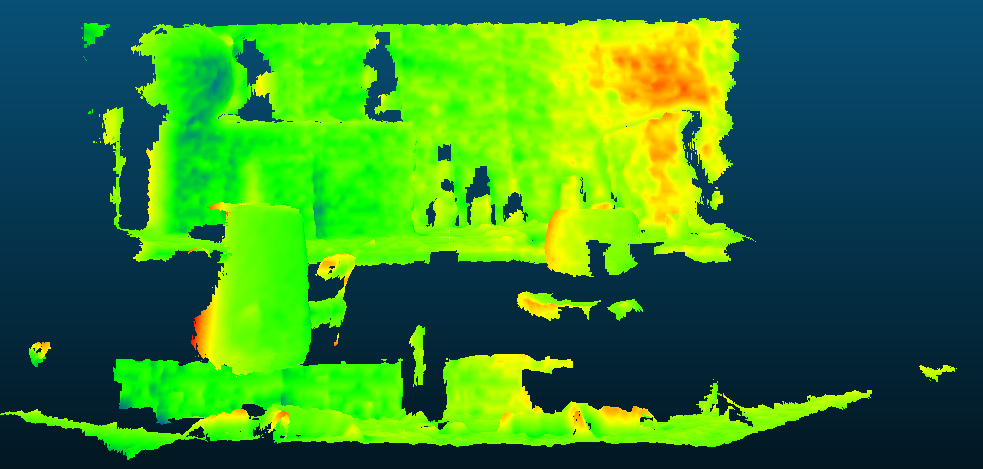}
	\includegraphics[width=0.07\linewidth]{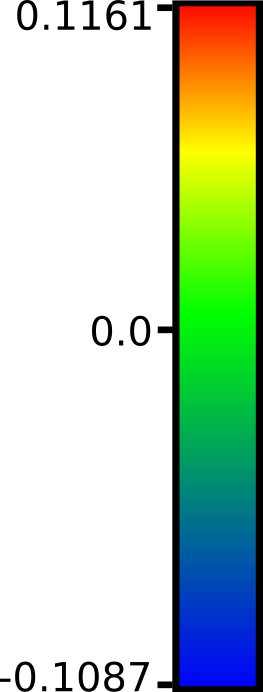}
	\caption[Reconstruction of $fr2-loh$ RGB-D data-set]{The reconstructed scene is based on $fr2-loh$ TUM collection. The actual trajectory consist of a circular movement around a table with verity of vibrations and rotational motions.}
	\label{fig:res_office}
\end{figure}
\begin{figure}[h]
	\centering
	\includegraphics[width=0.42\linewidth]{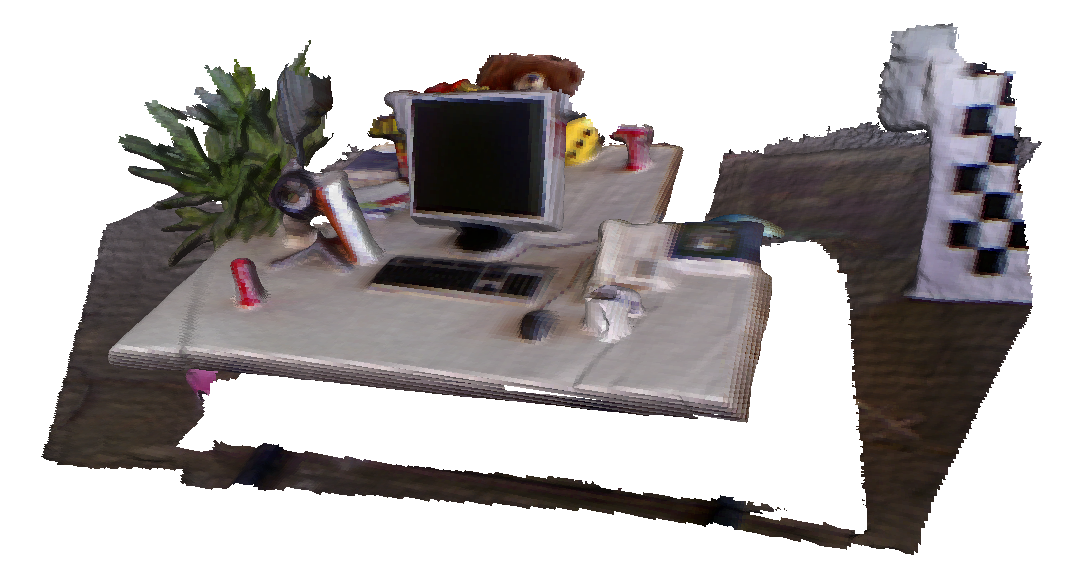}
	\includegraphics[width=0.42\linewidth]{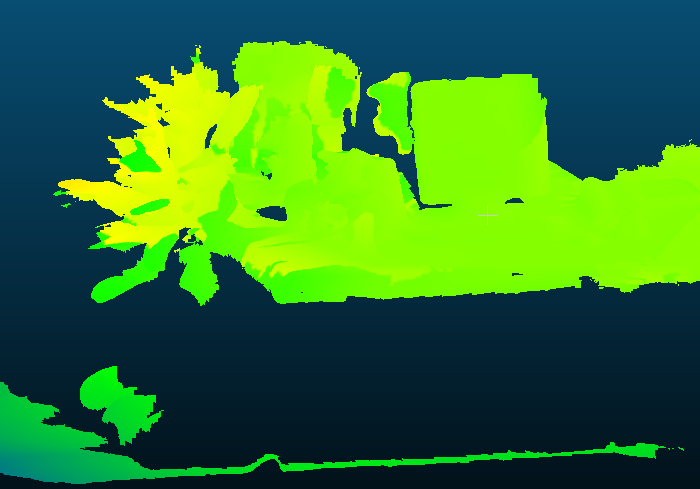}
	\includegraphics[width=0.07\linewidth]{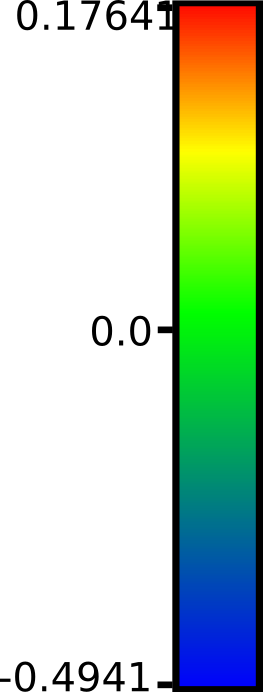}
	\caption[Reconstruction of fr2-xyz RGB-D data-set]{The reconstructed 3D RGB map is based on registration of more that 1000 frames from $fr2-xyz$ RGB-D data-set. The reconstructed surface is also evaluated w.r.t the reference model and boundaries if error values are shown in color gradient bar in the left.}
	\label{fig:res_fr2_xyz}
\end{figure}

\begin{figure}[h]
	\centering
	\includegraphics[width=0.42\linewidth]{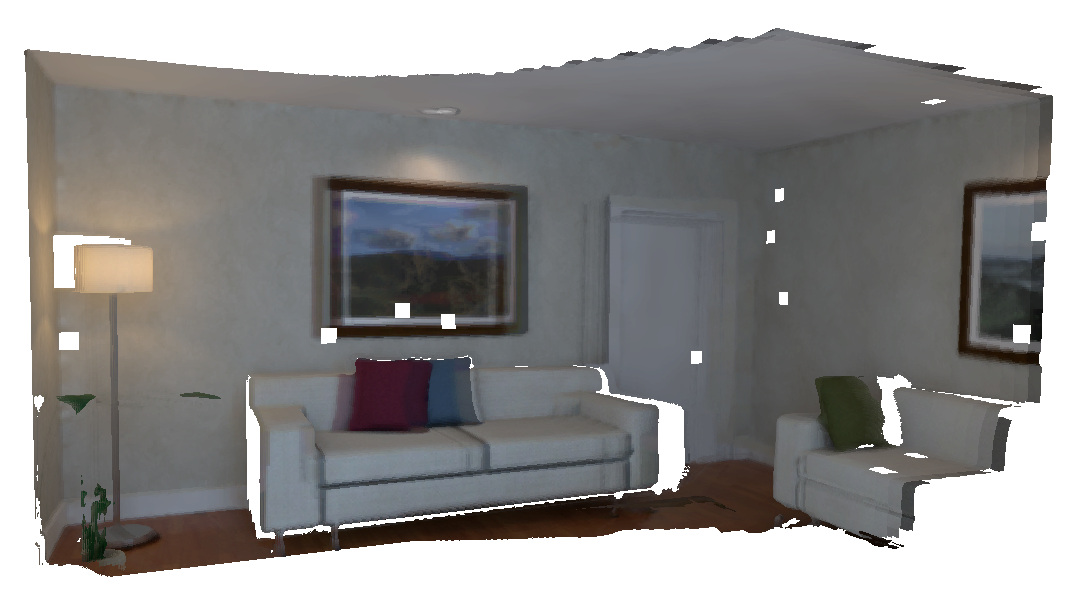}
	\includegraphics[width=0.42\linewidth]{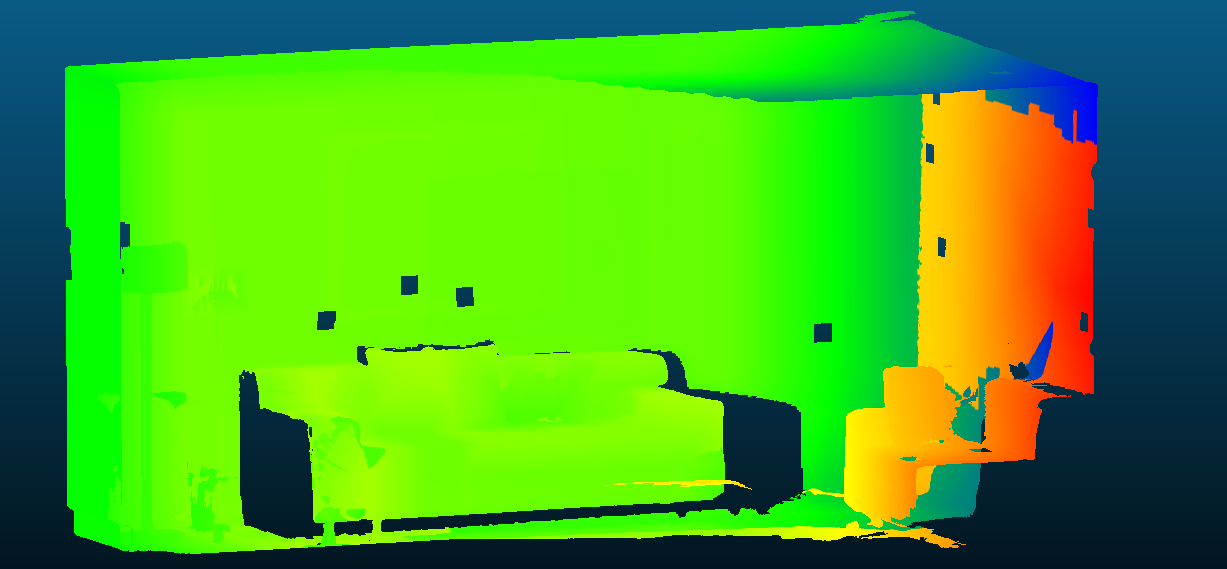}
	\includegraphics[width=0.07\linewidth]{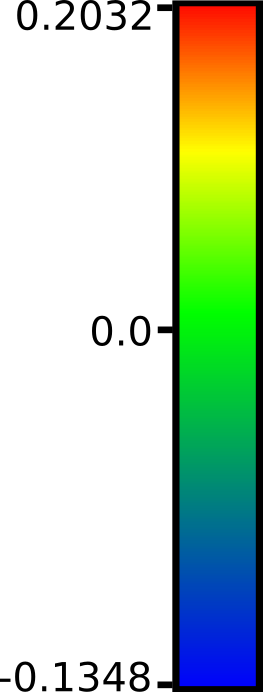}
	\caption[Reconstruction of $NUIM lv2-traj$ RGB-D data-set]{The results of 3D reconstruction of NUIM $lv2-traj$ virtual RGB-D data-set.}
	\label{fig:res_livingroom2}
\end{figure}

\begin{figure}[h]
	\centering
	\includegraphics[width=1.0\linewidth]{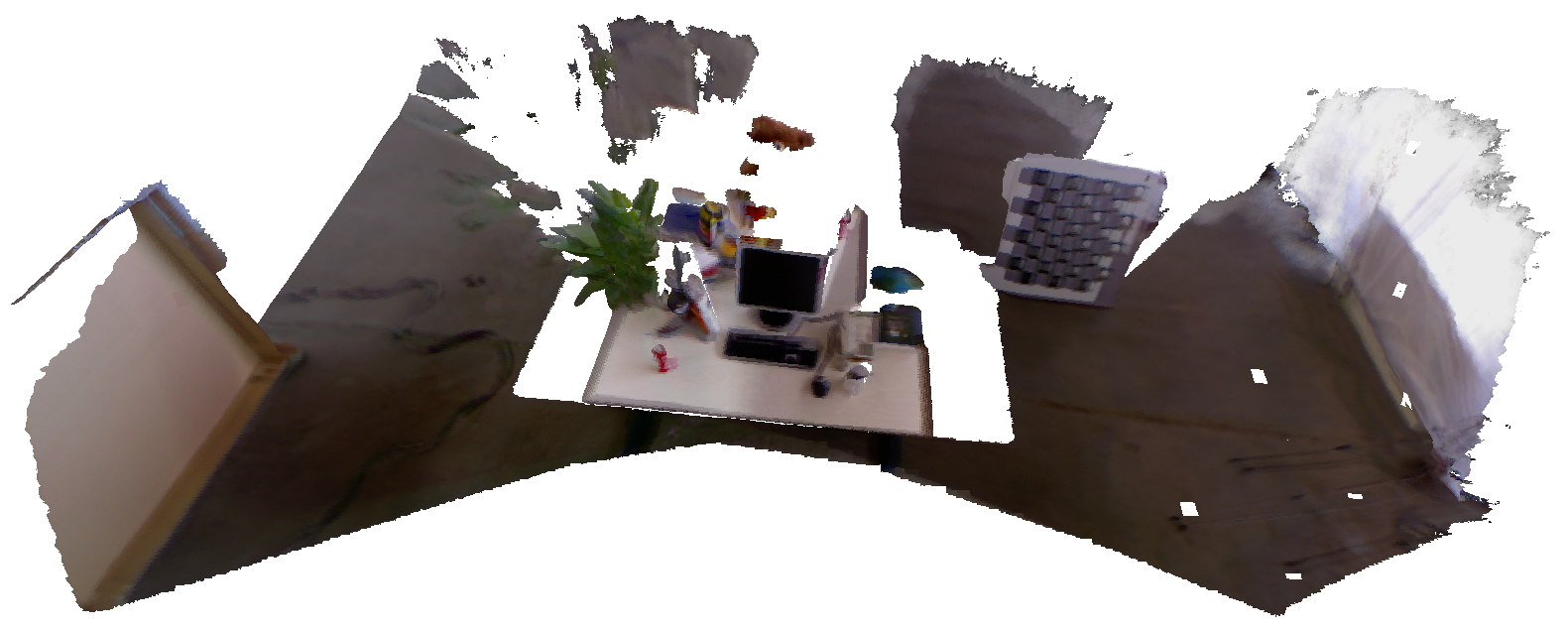}
	\caption[Reconstruction of $fr2-rpy$ RGB-D data-set]{Complete reconstruction of $fr2-rpy$ RGB-D data-set}
	\label{fig:res_fr2_rpy_comp}
\end{figure}

\cleardoublepage{}

\chapter{Dense Non-Rigid SLAM}
\label{chap:dense_non_rigid_slam}
% !TEX root = ../thesis.tex
The non-rigid SLAM is a crucial problem, as it provides a more realistic understanding of dynamic behaviors of the target. The main challenge in this area is that, the leading assumption of environment rigidity and linearity of the motions in the rigid mapping doesn't hold anymore. Also the sequential scans are partially overlapping with each other, and these regions of correspondence are not known priory \cite{li2008global}. This means the optimization problem defined in the section \ref{chap:dense_rigid_slam} will not handle the objects which undergo deformations while being scanned. The rigid SLAM topic has been deeply studied and many mature techniques have been introduced showing great performances. But, handling 3D non-rigid sequential depth maps in a dynamic environment or at least taken from a non-rigid object is more challenging and is an area of active research.

For instance, if in a rigid SLAM problem scans from a rigid object or a static environment can be matched via only six alignment parameters, in a non-rigid case number of unknowns would be $6 \times n$, where $n$ represents the number of sampled points in the non-rigid scan. Furthermore, as the problem here is highly non-linear, the solutions typically suffers from many ambiguities. The main reason of having bigger unknown space it that the registration algorithm needs to estimate the relative alignment parameters \eqref{eq:unknowns-non-rigid} between different correspondent pairs of point between target and source scans of the object over the time. The vector $\xi_i$  in equation below contains unknown parameters of $i_{th}$ point in the correspondence set.

\begin{equation}
\label{eq:unknowns-non-rigid}
\xi_i = 
\begin{bmatrix}
\alpha_i & \beta_i & \gamma_i & x_i & y_i & z_i
\end{bmatrix}
\end{equation} 

In this work, we try to address the non-rigid SLAM problem using deformation graph method introduced in \cite{sumner2007embedded}. We assume to have a working rigid SLAM pipeline which can register new depth images into a volumetric 3D space and extract 3D surface of the canonical model as well. Then, reconstructed surface $\mathcal{S}_c$ will be ray cast-ed at the latest camera position $T_{vl}$, providing us a predicted depth map $\hat{\mathcal{D}}$.
Afterwards the predicted depth map with certain graph information would be used to match to the new input depth map from the non-rigid object, which will be used in rest of the optimization procedure. 
This chapter is dedicated to introducing the problem of nonrigid surface registration, defining the optimization pipeline and procedure of warping deformations in a 3D volumetric representation. 

\section{Related Works}
\label{sec:non_rigid_related_works}
Non-rigid deformation tracking can estimate the dynamics in the target object whereas a conventional rigid SLAM pipeline may result in a inconsistent result at the same situation. 
Most of the researches in this field use offline approaches, and regularly use template based methods \cite{gall2008drift} or performance capture method \cite{starck2007surface} to estimate the geometric changes between reference model and scanned object, while only few of them estimate and warp deformations online while they are building the 3D model of the object \cite{newcombe2015dynamicfusion}\cite{innmann2016volumedeform}.

The literature on template based non-rigid tracking where combining ICP and reduced deformable model are discussed in  \cite{li2012temporally},\cite{brown2007global} and \cite{li20133d}. Furthermore, template based real-time non-rigid objects mapping with simple deformations is demonstrated \cite{zollhofer2014real}. 
But mapping deformable objects without using any template is still an open subject, while most of the template-free researches in this context has been done in offline manner \cite{dou20153d}, \cite{wang2016capturing}. 

There are few works in the subject of template-free non-rigid mapping that are able to perform in real-time. Newcombe \etal ~\cite{newcombe2015dynamicfusion}, proposed the first approach capable of reconstruction and tracking of deformable shapes based on RGB-D scans in real-time. They construct a volumetric representation of the scanned object and at the same time optimize for the models deformations using a hierarchy of deformation graphs \cite{sumner2007embedded}. Also, similar implementation is proposed in \cite{innmann2016volumedeform} where a featured based tracking is extending the $DynamiFusion$ approach, resulting more reliable tracking and reconstruction.

In this work we aim to build a template-free tracking and reconstruction pipeline for a non-rigid object. We use deformation graph method to estimate and warp the non-rigidity of the scanned object. Furthermore, the whole implementation of optimization process takes place on a GPU resulting in an online system which can reach to real-time performance via further GPU optimization.

\clearpage

\section{Non-Rigid Registration}
\label{sec:non_rigid_registration}
\begin{figure}[h]
	\centering
	\includegraphics[width=0.9\linewidth]{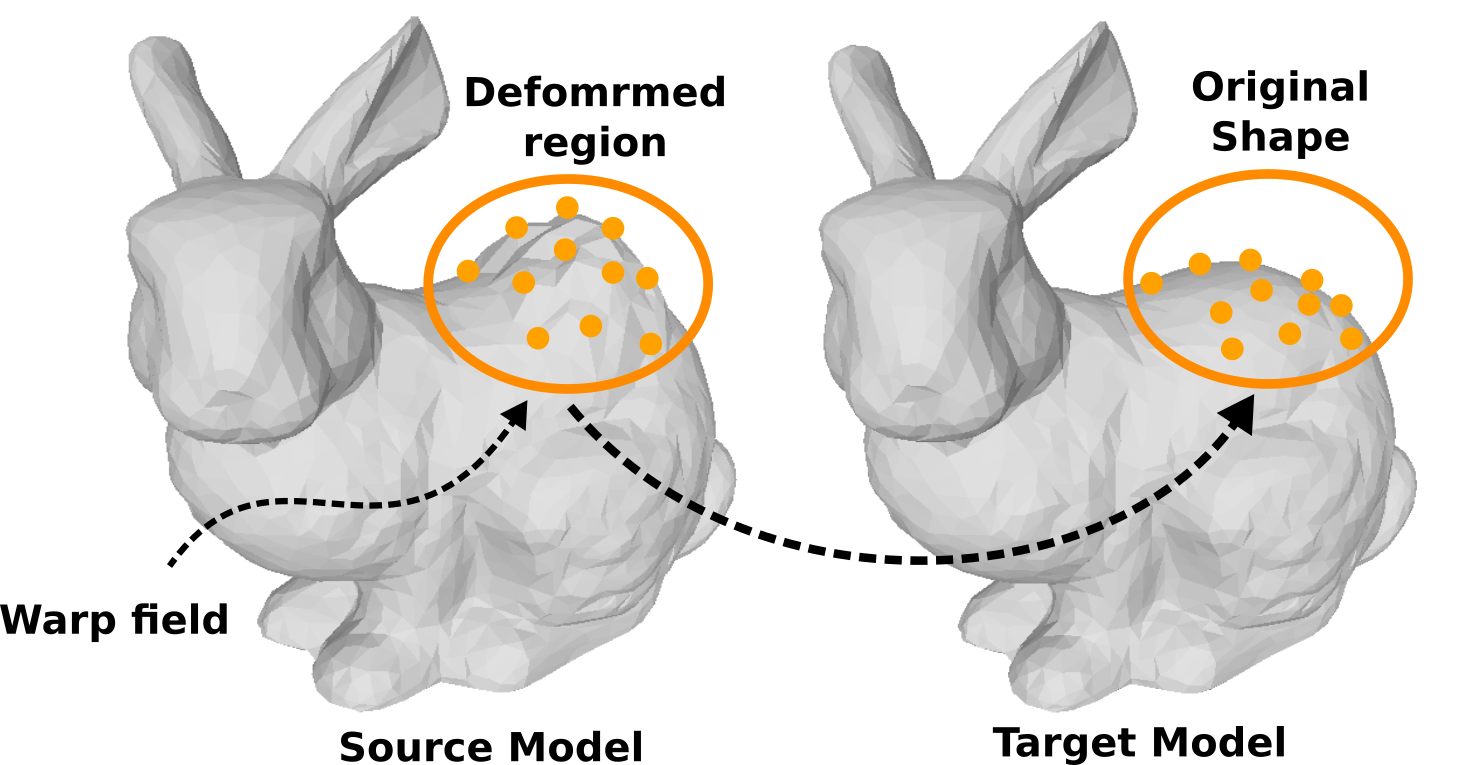}
	\caption[Non-rigid registration warp field initialization]{Sub-sampled points from target and source surfaces are assigned together as correspondences which initialize the warp field. This warp field state will be used in optimization procedure to retrieve original shape of the object.}
	\label{fig:non-reg-ex}
\end{figure}

We use a non-linear optimization method for non-rigid shapes that can find the optimal alignment parameters for partially matching scans \secref{sec:non_linear_least_squares}. The optimization procedure needs correspondences between source and target shapes and an initialized warp field, as it is shown in figure \figref{fig:non-reg-ex}

\begin{figure}[h]
	\centering
	\includegraphics[width=1\linewidth]{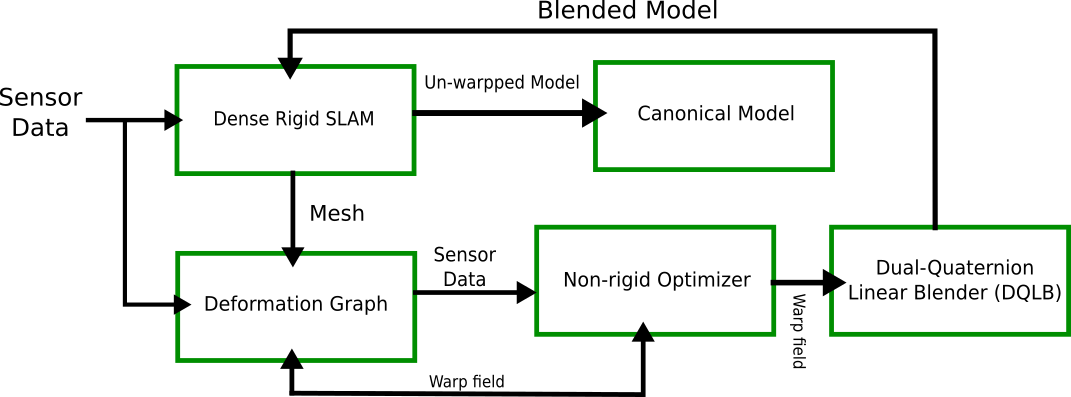}
	\caption[Non-rigid registration workflow]{Non-rigid registration workflow, New input scans get integrated intro the canonical model after estimation of their deformation space and un-warping them toward the canonical model.}
	\label{fig:non_rr_flowchart}
\end{figure}

Given the requirements, it will iteratively maximize the overlaying regions between source and target geometry, while reducing the registration error. As the main pipeline is inferred from ICP algorithm \ref{sec:camera_pose_estimation}, having a proper initial guess will accelerate the whole process and avoids getting stuck in a local minima. In figure \ref{fig:non_rr_flowchart} work flow of the non-rigid registration is depicted \cite{newcombe2015dynamicfusion} \cite{sumner2007embedded}.

The current canonical model $\mathcal{V}_c$ is a point-normal set stored as polygon mesh extracted from current zero level set of the $\mathbf{TSDF}$.

\begin{equation}
\label{eq:warpfunction}
\begin{split}
\mathcal{W}_t(\mathbf{v}_x, \mathbf{n}_x) & = \mathbf{DQLB}(\mathbf{v}_x, \mathbf{n}_x) \\
\mathcal{V}_w & = \mathcal{W}_t(\mathcal{V}_c)
\end{split}
\end{equation}

where, $\mathcal{V}_w$ denotes to the warped point-normal set and $\mathbf{DQLB}(\mathbf{v}_x, \mathbf{n}_x)$ is the Dual-Quaternion linear blender which is explained in section \ref{sec:dual_quat_linear_blending}. 

The registration pipeline needs a deformation blender which can apply estimated rigid body transformations on individual nodes and their surrounding vertices in each iteration. This function warps both position $\mathbf{v}_c$ and the normal $\mathbf{n}_c$ of each vertex in the canonical model $\mathcal{V}_c \equiv \{\mathbf{v}_c, \mathbf{n}_c\}$ towards the live frame's depth map $\mathcal{D}_t$ using latest estimate of warp field $\mathcal{W}_t$ according to equation \ref{eq:warpfunction}. 

\section{Deformation Graph}
\label{sec:deformation_graph}
\begin{figure}[h]
	\centering
	\includegraphics[width=0.5\linewidth]{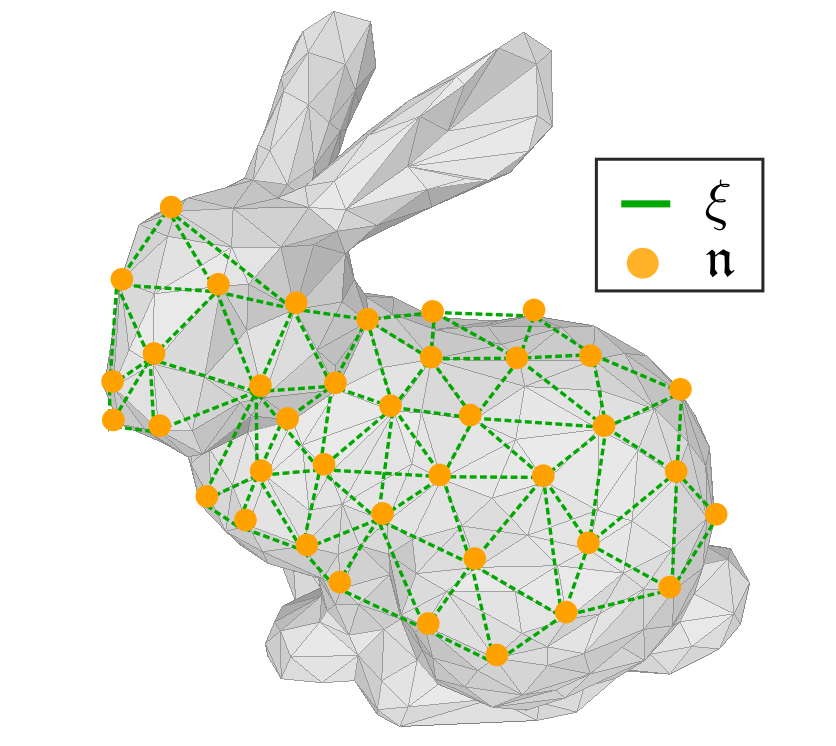}
	\caption[Warp filed initialization]{The warp field $\mathcal{W}_t$ is initialization with $n$ vertices of a mesh randomly, sub-sampling the mesh vertices.}
	\label{fig:dg_intro}
\end{figure}

We need to build a deformation graph to maintain the deformation information of the scanned object. We form a graph based on polygon mesh representation of current model which is called deformation graph. 
Sumner \etal, proposed a method to generate natural and intuitive deformations via using a triangular graph topology called deformation graph. This graph contains a series of rigid body transformations assigned to graph nodes $\mathfrak{n}$. Also, neighbor nodes in this structure share connecting edges $\xi$ along with a weight which defines the extent of influence which near nodes can have on each other according to figure \ref{fig:dg_intro}. We define the warp field $\mathcal{W}_t \equiv \{ \mathfrak{n}, \xi \}$ with $\mathfrak{n}$ as nodes and $\xi$ denoting internal edges between nodes as it is shown in \figref{fig:dg_intro}. 

By utilizing this method, we warp the deformations of two different surfaces toward each other. In this work, we use a general type of deformation graph which is enough to warp deformations of models with limited deformations while maintaining surface continuity. 

\subsection{Graph Construction}
\label{subsec:graph_construction}

First, we define structure of a single node as $\mathfrak{n}_i = \{ \mathbf{v}_i, \mathbf{w}_i, \mathbf{dq}_i \}$, where it has a position $\mathbf{v}_i$ in canonical model, its associated dual-quaternion $\mathbf{dq}_i$ equivalent to $4 \times 4$ transformation matrix and a weight $\mathbf{w}_i$ which defines the extent of the effecting nodes around that specific node according to equation \ref{eq:dq_weight}. We use dual-quaternions as a representation of transformation matrix associated with each node, which is used to apply the local deformations on the model vi $\mathbf{DQLB}$ method \secref{sec:dual_quat_linear_blending}.
We consider the total number of graph nodes to be $n$ with $K = 4$ nearest neighbors. 

\begin{equation}
\label{eq:dq_weight}
\mathbf{w}_i(x_k) = \exp \left(\dfrac{\lVert \mathbf{v}_i - x_k \rVert^2}{2 * \mathbf{w}_i^2} \right)
\end{equation}

where, $x_k$ denotes an arbitrary point in canonical model $\mathcal{V}_c $ near to $i_{th}$ node. Furthermore, the initial value of dual-quaternions in each node is set to identity and weights are set in such a way to ensure neighbor nodes share a proper overlapping area.

We uniformly distribute graph nodes on the polygon mesh $\mathcal{S}_c$ extracted from current canonical volumetric model $\mathcal{V}_c$, where $n$ vertices of the mesh will be randomly assigned as graph nodes. Next step would involve building a KD-tree from deformation graph nodes which can speed up accessing different nodes and their neighborhood information.

\section{Warp Field Optimizer}
\label{sec:warp_field_optimizer}

The main goal of this part is to estimate all rigid transformation parameters of all graph nodes. The estimated parameters should warp the canonical model $\mathcal{V}_c$ into the live frame depth map $\mathcal{D}_t$. Therefore, the registration is a model-to-frame optimization \cite{newcombe2015dynamicfusion}.

\begin{figure}[h]
	\centering
	\includegraphics[width=0.9\linewidth]{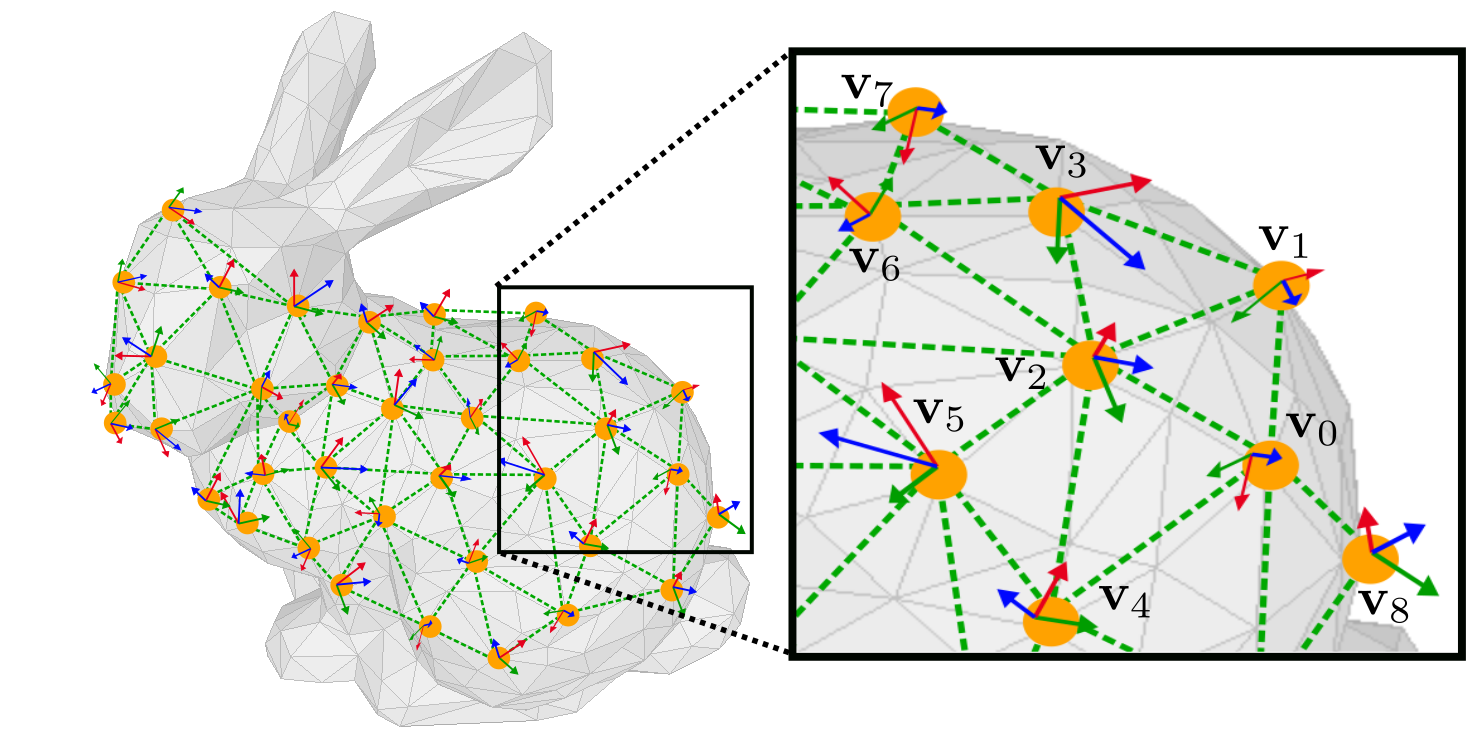}
	\caption[Warp field optimization]{The warp field optimization finds the best configuration of nodes transformation parameters to maximize the overlap between new scanned surface and wrapped canonical model.}
	\label{fig:warpfield_opt}
\end{figure}

The optimizer function updates the warp field $\mathcal{W}_t$ whenever a new depth map $\mathcal{D}_t$ is available given the current state of the canonical volume $\mathcal{V}_c$ by minimizing an energy function. We use a cost function containing the two following terms as:

\subsection{Data Term}
\label{subsec:data_term}
The data term is defined for direct association of two nodes in source and target point set, where the source vertex should transform to target vertex position after minimizing a squared distance error.
This cost term will maximize the overall similarity and overlap between live surface $\mathcal{V}_{vl}$ and warped canonical model $\mathcal{V}_w \equiv \{ \mathbf{v}_u, \mathbf{n}_u\}$. To address this problem, we need to first build the data association, which is done by ray casting $\mathcal{V}_w$ as a predicted depth map where, $\{ \mathbf{v}_c, \mathbf{n}_c\}: \Omega \longmapsto \mathcal{P}(\mathcal{V}_w)$ and associating it with current input depth map's back-projected vertices $\mathbf{vl}:\Omega \longmapsto \mathbb{R}^3$ \cite{newcombe2015dynamicfusion}, where the $\Omega$ is the pixel domain of the images. This can be quantified with a point-to-plane error metric as follows:

\begin{equation}
\label{eq:data_term}
\mathbf{E}_{Data}(\mathcal{W}_t, \mathcal{V}_w, \mathcal{D}_t) \equiv \sum_{u \in \Omega} \left( \mathbf{n}_u^\top (\mathbf{v}_u - \mathbf{vl}_u) \right)
\end{equation}

\begin{figure}[h]
	\centering
	\includegraphics[width=0.9\linewidth]{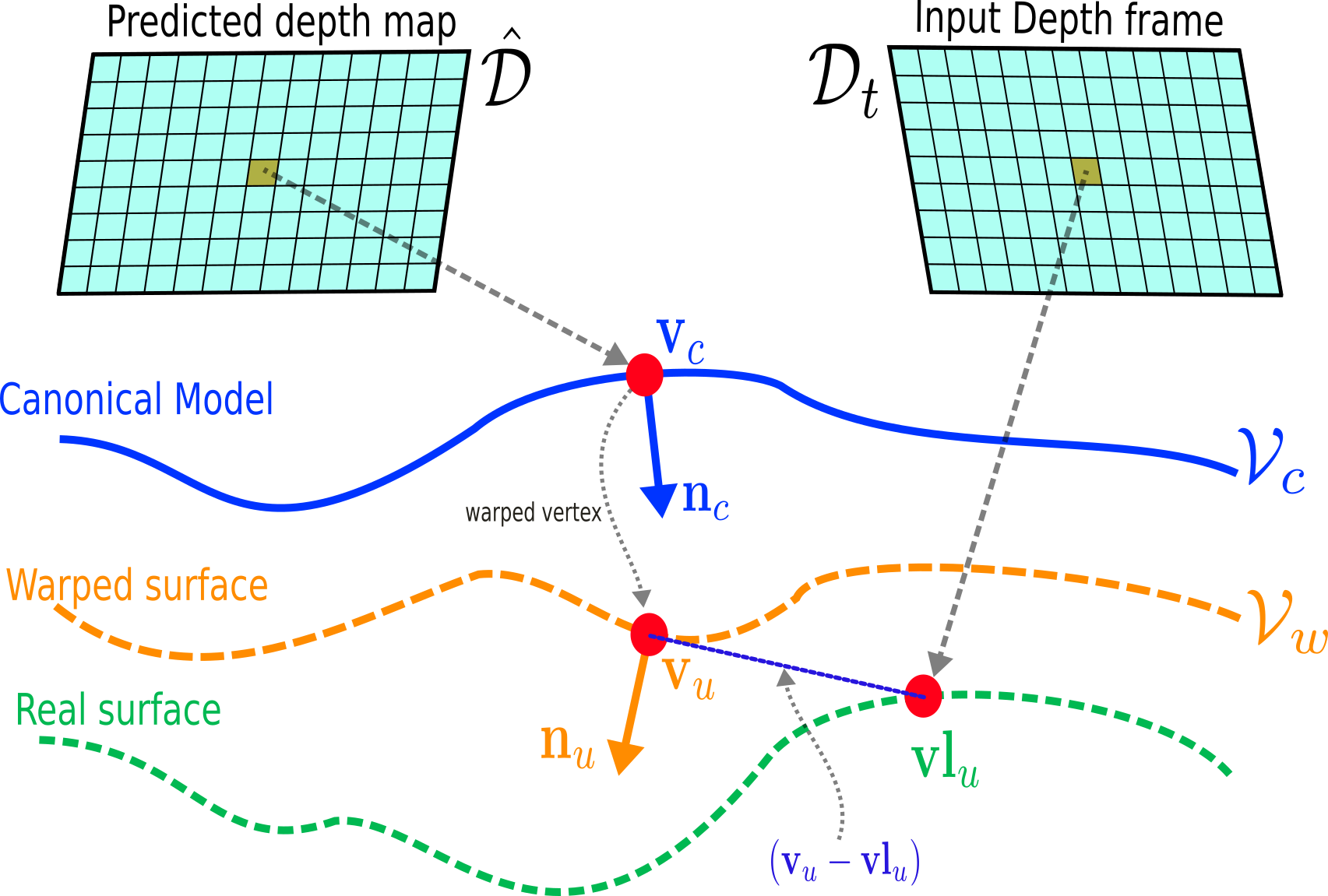}
	\caption[Data term]{The data term in total non-rigid optimization error minimizes distance between nodes in such a way to increase the overlap of target and source models. Here, vertex and normal $\mathbf{v}_c, \mathbf{n}_c$ are warped with latest warp field state. The warped surface is used in optimization process where it is ray cast-ed into image $\hat{\mathcal{D}}$ to be compared with input image $\mathcal{D}_t$. }
	\label{fig:warpfield_data}
\end{figure}

\subsection{Regularization Term}
\label{subsec:regularization_term}

The regularization cost will penalize any inconsistent motion between a node and its neighbors. thereby avoiding non-smooth and harsh deformations in a graph where each node is surrounded with $\mathbf{K}$ nearest neighbors. The regularization term assigns a cost between node $i$ and node $j$ which are connected by an edge \cite{newcombe2015dynamicfusion}. This cost term will keep neighbor nodes deformations consistent and can be defined  as:

\begin{equation}
\label{eq:reg_term}
\mathbf{E}_{Reg}(\mathcal{W}_t) \equiv \sum_{i = 0}^{n} \sum_{j \in \xi(i)} \alpha_{ij} (\mathbf{T}_{ic} \mathbf{v}_j - \mathbf{T}_{jc} \mathbf{v}_j)
\end{equation}

where $\alpha_{ij}$ defines the weight associated with edge according to:

\begin{equation}
\label{eq:reg_alpha}
\alpha_{ij} = \max(\mathbf{w}_i, \mathbf{w}_j)
\end{equation}
 
\begin{figure}[h]
	\centering
	\includegraphics[width=0.8\linewidth]{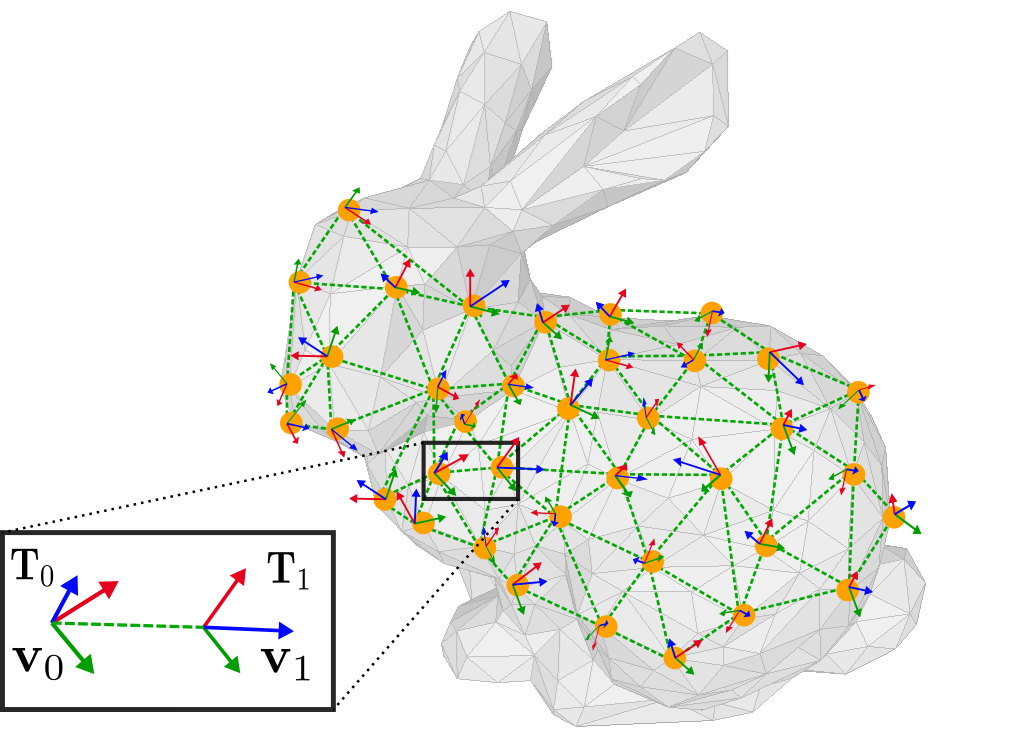}
	\caption[Regularization term]{The regularization term penalizes harsh deformations and movements in the process of non-rigid optimization. This is done by setting a cost based on neighbor nodes poses. In such a way that the transformation of each node $i$ should be consistent with its warped neighbor nodes' $j$ position after applying estimated deformation parameters.}
	\label{fig:warpfield_reg}
\end{figure}

The equation \ref{eq:non_r_error} shows the total cost function used in this work which is a weighted sum of all mentioned cost terms.

\begin{equation}
\label{eq:non_r_error}
\mathbf{E} (\mathcal{W}_t, \mathcal{V}_w, \mathcal{D}_t, \varphi) = \mathbf{E}_{Data}(\mathcal{W}_t, \mathcal{V}_w, \mathcal{D}_t) +  \varphi \mathbf{E}_{Reg}(\mathcal{W}_t)
\end{equation}

where, the regularization term is controlled by weight parameter $\varphi$ and will ensure having a smooth and consistent motion using as-rigid-as-possible axiom introduced by Sorkine \etal \cite{sorkine2007rigid}. 
 
We estimate the unknowns by linearizing total cost function \ref{eq:non_r_error} around the current estimate of graph node transformations. This we will form the normal equations \ref{eq:nr_gauss} and by taking steps toward the local descent direction will results in minimizing the cost function, as Gauss-Newton non-linear optimization approach suggests \secref{subsec:gauss_newton_method}. 

\begin{equation}
\label{eq:nr_gauss}
\mathbf{J}^\top \mathbf{J} \mathbf{\hat{x}} = \mathbf{J}^\top \mathbf{r}
\end{equation}

where $\mathbf{\hat{x}}$ defines the unknowns vector including $6 \times n$ nodes alignment parameters. The size of the problem enforces us to utilize an  efficient solver and problem building procedure, because the deformation graph may contain hundreds of nodes, which leads to have gigantic Jacobian matrices. Further more constructing and factorizing Jacobian matrices are computationally expensive, as well. Therefore we needs to accelerate the whole process via GPU multi-threading techniques. As long as the main cost function \ref{eq:non_r_error} consists of two different parts, we build up the linear system separately for each term according to: 

\begin{equation}
\label{eq:nr_normal}
\begin{split}
\mathbf{J}^\top \mathbf{J} & = \mathbf{J}_{Data}^\top \mathbf{J}_{Data} + \varphi \mathbf{J}_{Reg}^\top \mathbf{J}_{Reg} \\[8pt]
\mathbf{J}^\top \mathbf{r} & = \mathbf{J}_{Data}^\top \mathbf{r}_{Data} + \varphi \mathbf{J}_{Reg}^\top \mathbf{r}_{Reg}
\end{split}
\end{equation}

Whereas the size of this optimization is really big, the real-time performance of solvers like Cholesky factorization will not be hindered due to the sparsity of the matrices. We can take a closer look at normal equations and procedure of creating Jacobian matrices. The matrix $\mathbf{J}_{Data}$ is containing $6n$ columns and $m$ rows, where $m$ shows number of observation.

\begin{equation}
\label{eq:j_nr_data}
Block_{ij} \equiv 
\begin{bmatrix}
\frac{\partial \mathbf{E}_{data}}{\partial \alpha_j} & 
\frac{\partial \mathbf{E}_{data}}{\partial \beta_j} & 
\frac{\partial \mathbf{E}_{data}}{\partial \gamma_j} & 
\frac{\partial \mathbf{E}_{data}}{\partial x_j} & 
\frac{\partial \mathbf{E}_{data}}{\partial y_j} & 
\frac{\partial \mathbf{E}_{data}}{\partial z_j}
\end{bmatrix}
\end{equation}  

where each row contains derivatives of the $E_{data}$ term w.r.t the $K$ nearest nodes which are affecting the position of the corresponding vertex. Each block of 6 elements belongs to one node and only $K$ blocks are non-zero in each row, resulting a sparse matrix. The arrangements of the elements are shown in \eqref{eq:j_data}.

\begin{equation}
\label{eq:j_data}
\mathbf{J}_{Data} \equiv 
\begin{bmatrix}
0 & 0 & 0 & B_{00} & B_{03}  & \dots & B_{01} & B_{02} & 0 & 0\\
B_{13} & B_{11} & 0 & 0 & B_{10}  & \dots & 0 & 0 & 0 & B_{12}\\
\vdots & \vdots & \vdots & \vdots &  \vdots & \vdots & \vdots & \vdots &\vdots  &\vdots \\
\vdots &    &   &    & &  & & & & \\
\vdots &    &   &    & & &  & & & \\
\vdots &    &   &    & & & &  & & \\
\vdots &    &   &    & & & & &  & \\
0 & 0 & B_{n3} & 0 & B_{n0} & \dots & 0 & B_{n2} & 0 & B_{n1}\\
\end{bmatrix}_{m \times 6n}
\end{equation}

The $\mathbf{J}_{Reg}$ is involved with computing normal equations and filling a matrix with size of $3m \times 6n$. The elements of this matrix are grouped in blocks with size of $3 \times 6$. Each elements is assigned according to \ref{eq:nr_reg_mat} and the whole matrix scheme is shown in \eqref{eq:j_reg}.
\begin{equation}
\label{eq:reg_term_e}
\mathbf{E}_{ij} = (
\underbrace{
\mathbf{T}_{ic} \mathbf{v}_j}_{a_{4 \times 1}}-
\underbrace{
\mathbf{T}_{jc} \mathbf{v}_j)}_{b_{4 \times 1}}
\end{equation}

where the $a_{ij}$ and $b_{ij}$ are defined as following:

\begin{equation}
\label{eq:j_nr_reg}
a_{ij} = 
\begin{pmatrix}
1 & -\gamma_i & \beta_i & tx_i	\\
\gamma_i & 1 & -\alpha_i & ty_i	\\
-\beta_i & \alpha_i & 1 & tz_i	\\
0   & 0   & 0  & 1
\end{pmatrix}
\begin{pmatrix}
x_j	\\
y_j	\\
z_j	\\
1
\end{pmatrix}
\end{equation}

\begin{equation}
\label{eq:j_nr_reg_1}
b_{ij} = 
\begin{pmatrix}
1 & -\gamma_j & \beta_j & tx_j	\\
\gamma_j & 1 & -\alpha_j & ty_j	\\
-\beta_j & \alpha_j & 1 & tz_j	\\
0   & 0   & 0  & 1
\end{pmatrix}
\begin{pmatrix}
x_j	\\
y_j	\\
z_j	\\
1
\end{pmatrix}
\end{equation}

by taking the derivative of the $a_{ij}$ and $b_{ij}$ w.r.t $\xi_i$ we will get:

\begin{equation}
\label{eq:nr_reg_mat}
\begin{split}
Block_{ii} \equiv \frac{\partial \mathbf{E}_{a}}{\partial \xi_i} & = 
\begin{bmatrix}
0 & z_i & - y_i & 1 & 0 & 0 \\
-z_i & 0 &  x_i & 0 & 1 & 0 \\
y_i & - x_i &  0 & 0 & 0 & 1
\end{bmatrix} \\
Block_{ij} \equiv \frac{\partial \mathbf{E}_{b}}{\partial \xi_j} & = 
\begin{bmatrix}
0 & z_j & - y_j & 1 & 0 & 0 \\
-z_j & 0 &  x_j & 0 & 1 & 0 \\
y_j & - x_j &  0 & 0 & 0 & 1
\end{bmatrix}
\end{split}
\end{equation}

and the overall scheme of Jacobian matrix $\mathbf{J}_{Reg}$ is shown in below:

\begin{equation}
\label{eq:j_reg}
\mathbf{J}_{Reg} \equiv 
\begin{bmatrix}
B_{00} & 0 & 0 & B_{n0} & B_{03}  & \dots & B_{01} & B_{02} & 0 & 0\\
B_{13} & B_{11} & 0 & 0 & B_{10}  & \dots & 0 & 0 & 0 & B_{12}\\
\vdots & \vdots & \ddots & \vdots &  \vdots & \vdots & \vdots & \vdots &\vdots  &\vdots \\
\vdots &    &   &    & &  & & & & \\
\vdots &    &   &    & & &  & & & \\
\vdots &    &   &    & & & &  & & \\
\vdots &    &   &    & & & & &  & \\
0 & B_{n0} & B_{n3} & 0 & B_{nn} & \dots & 0 & B_{n2} & 0 & B_{n1}\\
\end{bmatrix}_{3m \times 6n}
\end{equation}

 \clearpage 
 
\section{Results and Evaluation}
\label{sec:nr_results_and_Evaluation}
\begin{figure}[h]
	\centering
	\includegraphics[width=0.35\linewidth]{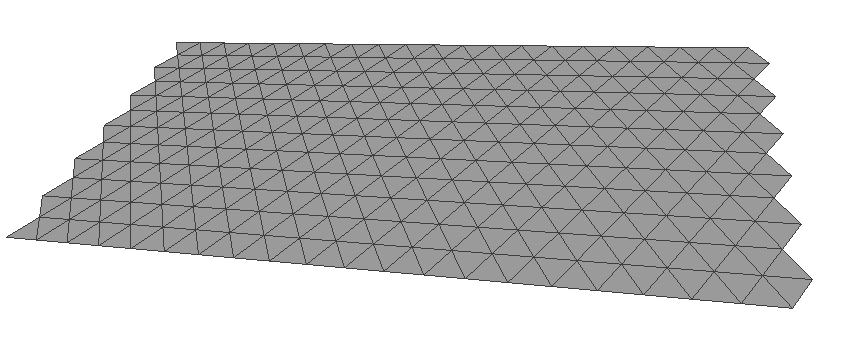}
	\caption[Reference flat plane mesh]{A planar mesh with 273 vertices and 480 triangles used as source model of non-rigid optimizer.}
	\label{fig:ref_plane}
\end{figure}

We demonstrate the performance of our method with a series of experiments on synthetic test data. Our test includes evaluating the performance of proposed method on synthetic data where the ground truth is given. During this test, the non-rigid optimizer gets two different planar meshes as input. Then the algorithm should warp the first mesh toward the second one in such a way to minimize total cost \eqref{eq:non_r_error}. In our test the source mesh is always a flat plane with 273 vertices and 480 triangles shown in \figref{fig:ref_plane}. The results of this test on different types of deformations are shown in table \tabref{tab:nr_stat_synthetic_res} and \tabref{tab:nr_synthetic_res}. 

The statistical results from table \tabref{tab:nr_stat_synthetic_res} show the performance of our approach in estimating and warping deformation of two different planes with different types of deformations, shown in \tabref{tab:nr_synthetic_res}. In this implementation, we do not use any hierarchical sub-sampling technique between source and target shapes. As a consequence in case of having a large difference in surfaces, the algorithm will not be able to warp the deformations optimally. Moreover, tuning the parameters like the number of graph nodes, the number of considered neighbors for each node in the graph and the trade-off parameter between cost functions $\varphi$ play a significant role in getting proper results. For example, in cases 4 and 5 the source and the target planes are same, but in case 4 having a big $\varphi$ caused a smoother output while the registration error is not minimized satisfyingly. 

\begin{table}[h]
	\begin{center}
		\caption{Statistical results of synthetic evaluations of non-rigid optimizer, all values are .}
		\label{tab:nr_stat_synthetic_res}
		\begin{tabular}{l c c c c c c} % <-- Alignments: 1st column left, 2nd middle and 3rd right, with vertical lines in between
			& Max dis. & Max dis. ref. & max avg.  &  Max avg. ref. & $\sigma$ & $\sigma$ ref.\\
			\hline
			\hline
			1 & 4e-3    & 8.5e-3 & 1.64e-6 & 2.29e-6 & 6.9e-6 & 1.2e-4 \\
			2 & 8.5e-3  & 2.4e-2 & 1.21e-5 & 2.43e-5 & 2.6e-4 & 5.3e-4 \\
			3 & 1.3e-2  & 1.8e-2 & 7.83e-6 & 7.81e-2 & 2.1e-4 & 1.2e-2 \\
			4 & 2.3e-2  & 3.9e-2 & 1.92e-5 & 3.25e-5 & 4.4e-4 & 8e-4   \\
			5 & 2.4e-2  & 3.9e-2 & 1.67e-5 & 3.25e-5 & 4.1e-4 & 8e-4   \\
		\end{tabular}
	\end{center}
\end{table}

\begin{table}[h]
	\begin{center}
		\caption[Visual results of synthetic evaluations of non-rigid optimizer]{Visual results of synthetic evaluations of non-rigid optimizer. $left$: the target model of non-rigid optimizer, $middle$: color coded error visualization of target and source planes, $right$: the result of optimization and warping deformations.}
		\label{tab:nr_synthetic_res}
		\begin{tabular}{l c c c} % <-- Alignments: 1st column left, 2nd middle and 3rd right, with vertical lines in between
			& Target Model & Ground truth & Result \\
			\hline
			1&
			\includegraphics[width=0.3\linewidth]{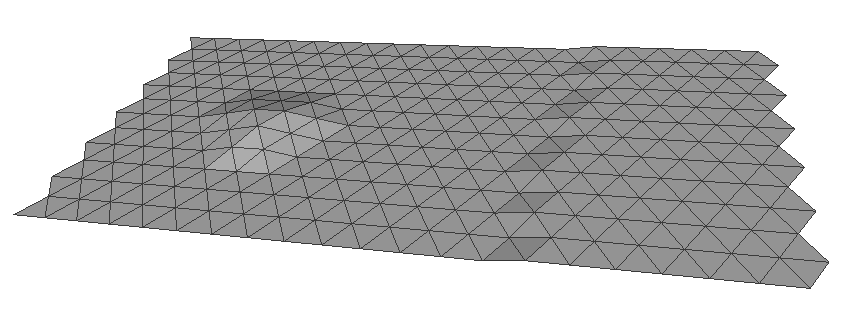}
			&
			\includegraphics[width=0.3\linewidth]{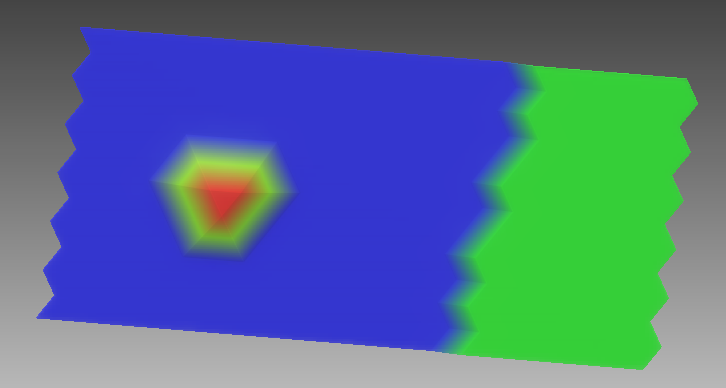}
			&
			\includegraphics[width=0.3\linewidth]{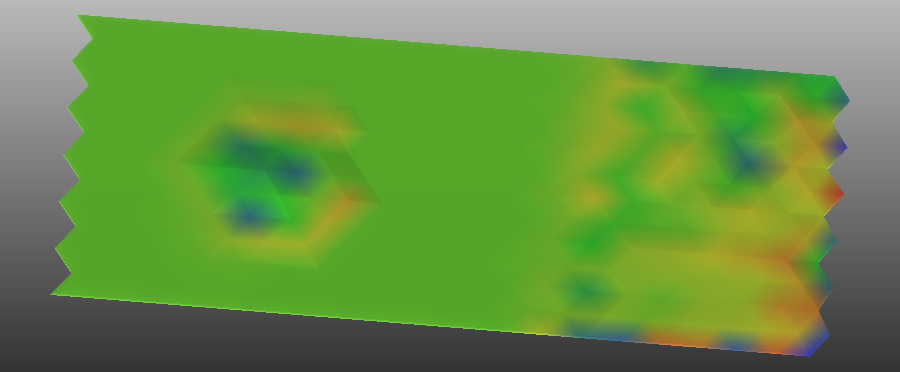}\\
			2&
			\includegraphics[width=0.3\linewidth]{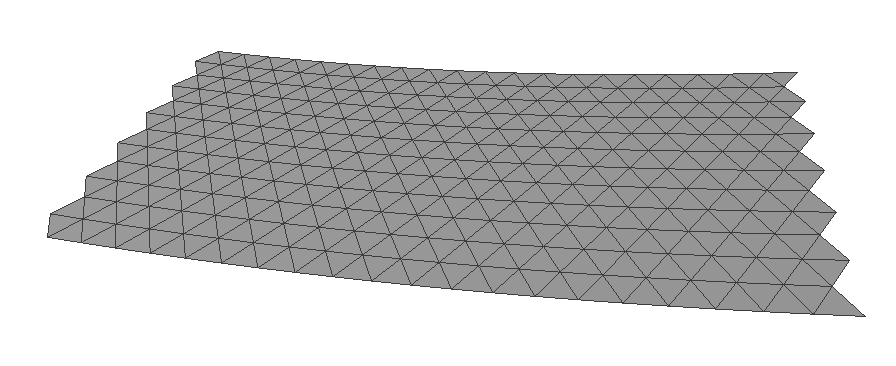}
			&
			\includegraphics[width=0.3\linewidth]{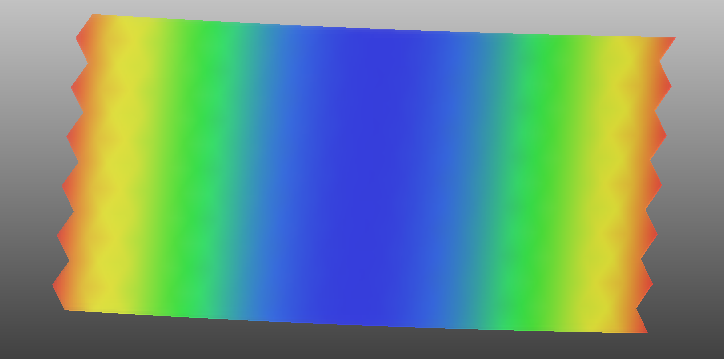}
			&
			\includegraphics[width=0.3\linewidth]{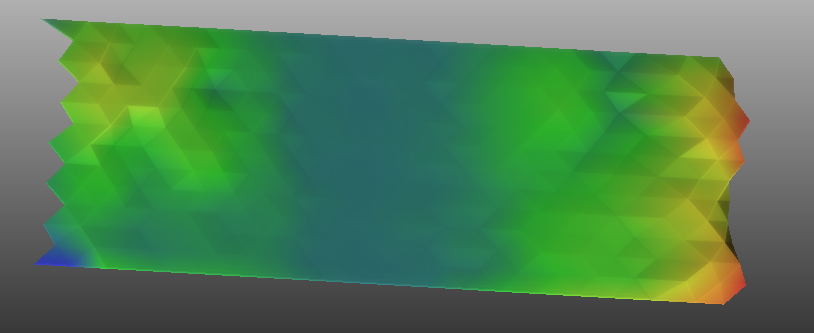}\\
			3&
			\includegraphics[width=0.3\linewidth]{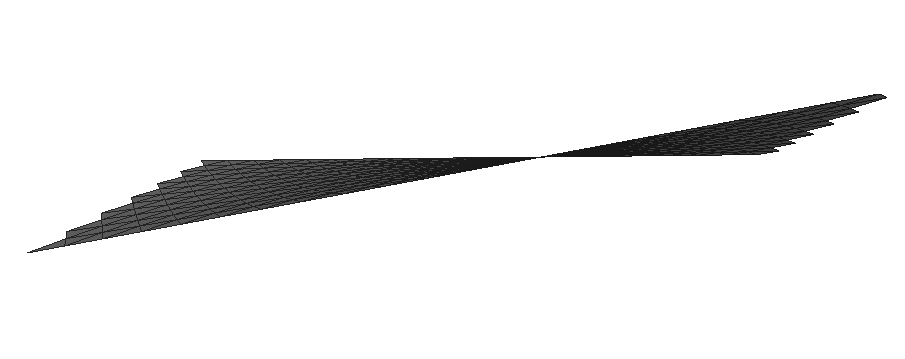}
			& 
			\includegraphics[width=0.3\linewidth]{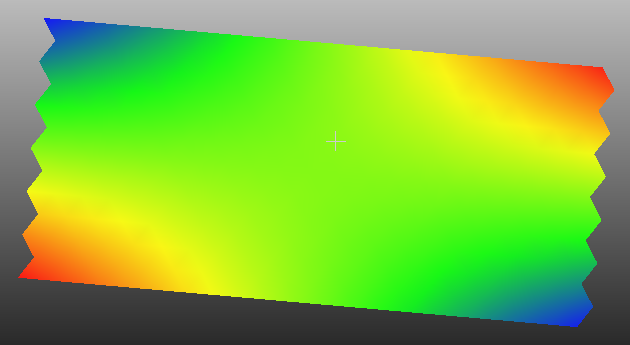}
			& 
			\includegraphics[width=0.3\linewidth]{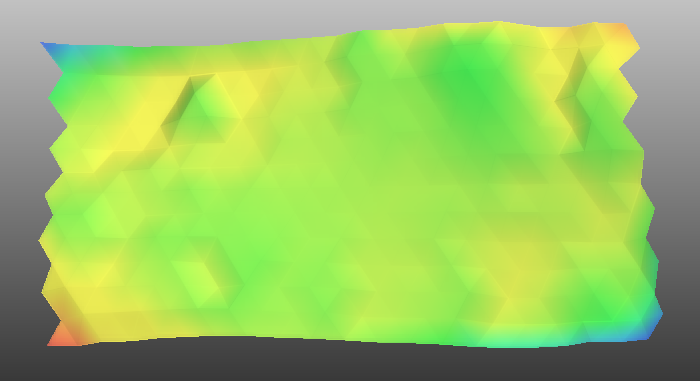}\\
			4&
		    \includegraphics[width=0.3\linewidth]{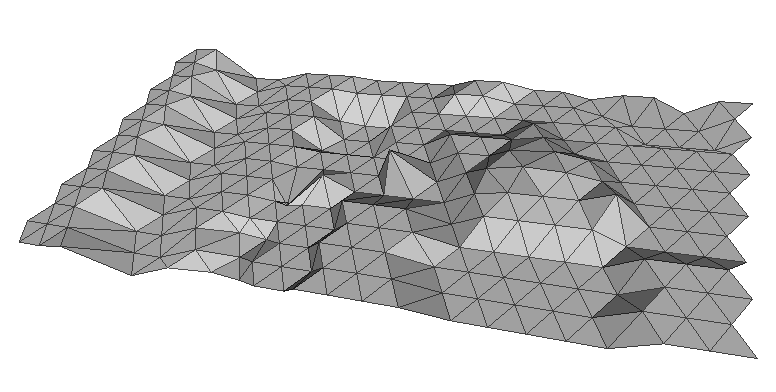} & \includegraphics[width=0.3\linewidth]{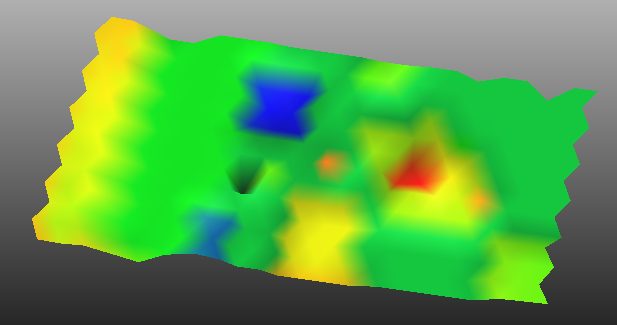} & \includegraphics[width=0.3\linewidth]{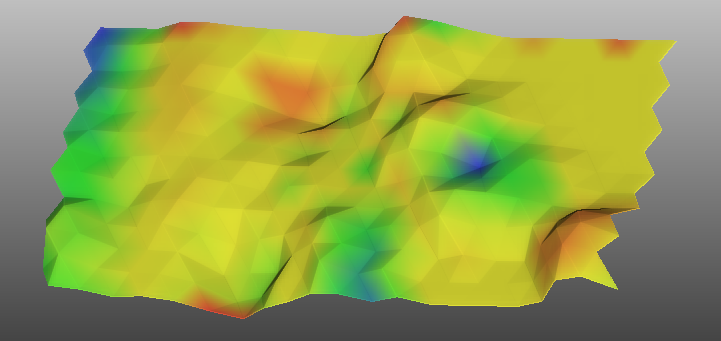} \\
		    5&
		    \includegraphics[width=0.3\linewidth]{pics/violent} & \includegraphics[width=0.3\linewidth]{pics/ref_violent} & \includegraphics[width=0.3\linewidth]{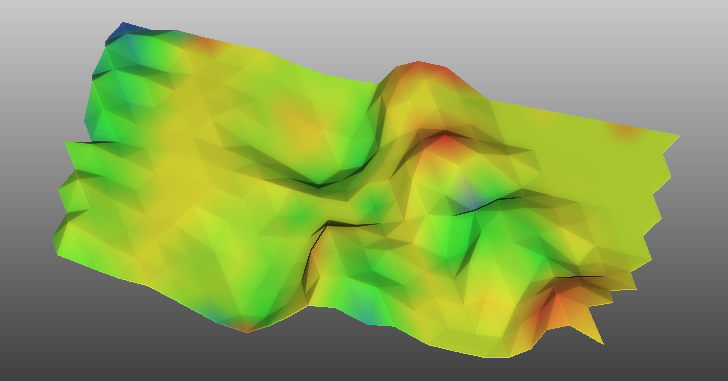} \\
		\end{tabular}
	\end{center}
\end{table}

\clearpage

\cleardoublepage{}

\bibliographystyle{plain} % We choose the "plain" reference style
\bibliography{references} % Entries are in the "refs.bib" file

\end{document}